\documentclass{article}
\usepackage[utf8]{inputenc}
\usepackage{microtype}
\usepackage{graphicx} 
\usepackage{booktabs} 
\usepackage{makecell}
\usepackage{longtable} 
\usepackage{subcaption}
\usepackage{tabularx}
\usepackage{multirow} 
  
\usepackage{pgfplotstable}
\usepgfplotslibrary{groupplots}
  
\usepackage[dvipsnames,table]{xcolor}
\usepackage{colortbl}

\usepackage{caption}
\usepackage{placeins} 
  
\usepackage{xurl}
\usepackage[numbers]{natbib}

\usepackage[preprint]{icml2026}
\usepackage{pgfmath}
\usepackage{amsmath}
\usepackage{amssymb}
\usepackage{mathtools}
\usepackage{amsthm} 
\usepackage{pgfplots}
\usepackage{tikz}
\usepackage{tcolorbox}
\usepackage{listings}
\usepackage{color}
\usepackage{fancybox}
\usepackage{mdframed}   
\usepackage{threeparttable} 
   
\usepackage{arydshln}

\definecolor{keywordcolor}{rgb}{0.7, 0.1, 0.1}   
\definecolor{tacticcolor}{rgb}{0.0, 0.1, 0.6}    
\definecolor{commentcolor}{rgb}{0.1, 0.5, 0.1}   
\definecolor{symbolcolor}{rgb}{0.0, 0.1, 0.6}    
\definecolor{sortcolor}{rgb}{0.1, 0.5, 0.1}      
\definecolor{attributecolor}{rgb}{0.7, 0.1, 0.1} 
\definecolor{springcolor}{rgb}{0, 0, 0}

\definecolor{orange}{cmyk}{0,0.5,1,0}   

\usepackage{hyperref}
\usepackage[capitalize,noabbrev]{cleveref}

\theoremstyle{plain}

\theoremstyle{definition}

\theoremstyle{remark}

\usepackage[textsize=tiny]{todonotes} 
\definecolor{bgcolor}{RGB}{245,245,245}
\definecolor{framecolor}{RGB}{200,200,200}
\definecolor{titlecolor}{RGB}{0,102,204}

\icmltitlerunning{Submission and Formatting Instructions for ICML 2026}

\begin{document}

\twocolumn[
\icmltitle{Automated Formal Proofs of Combinatorial Identities via Wilf–Zeilberger Guidance and LLMs}




\icmlsetsymbol{equal}{*}
\begin{icmlauthorlist}
\icmlauthor{Beibei Xiong}{ecnu}
\icmlauthor{Hangyu Lv}{ecnu}
\icmlauthor{Junqi Liu}{iss}
\icmlauthor{Yisen Wang}{iss}
\icmlauthor{Shaoshi Chen}{iss}
\icmlauthor{Jianlin Wang}{henu}
\icmlauthor{Zhengfeng Yang}{ecnu} 
\icmlauthor{Lihong Zhi}{iss}
\icmlcorrespondingauthor{Zhengfeng Yang}{zfyang@sei.ecnu.edu.cn}
\end{icmlauthorlist}

\icmlaffiliation{ecnu}{School of Software Engineering, East China Normal University, Shanghai, China}
\icmlaffiliation{henu}{School of Computer and Information Engineering, Henan University, Kaifeng, China}
\icmlaffiliation{iss}{Institute of Mathematics and Systems Science, Chinese Academy of Sciences, Beijing, China}

\icmlkeywords{Machine Learning, ICML}

\vskip 0.3in
]



\printAffiliationsAndNotice{}  

\begin{abstract}

Automating formal proofs of combinatorial identities is challenging for LLM-based provers, as long-horizon proof planning is required and unconstrained search quickly explodes. 
Symbolic methods such as the Wilf--Zeilberger (WZ) method can achieve a mechanized proof of combinatorial identities by constructing special auxiliary functions and demonstrating that they satisfy specific recurrence relations. 
We propose WZ-LLM, a neuro-symbolic framework that turns WZ proof plans into \emph{executable proof sketches} in Lean~4 and uses an LLM-based prover 
to discharge the resulting machine-checkable subgoals.
We also train a dedicated WZ-Prover via a Lean-kernel-verified bootstrapping loop with expert-verified iteration, followed by DAPO-based refinement.
Experiments show that WZ-LLM achieves a 34\% proof success rate on LCI-Test (100 classic combinatorial identities), outperforming strong baselines such as DeepSeek-V3 and Goedel-Prover-V2, and delivering consistent gains on CombiBench and PutnamBench-Comb. These results indicate that our framework provides two complementary strengths: improved direct proving for identities beyond the scope of WZ, and substantially higher end-to-end success when WZ sketches guide a specialized prover.

\end{abstract}

\section{Introduction}

Large language models (LLMs) have recently achieved strong performance on formal automated theorem proving (ATP) benchmarks~\cite{xin2025bfs,lin2025goedel,guo2025deepseek,wang2025kimina,chen2025seedproverdeepbroadreasoning}, enabling formal reasoning toward research-level mathematics~\cite{wei2024proving,yu2025formalmath}. 
However, scaling LLM-based provers to harder mathematical domains remains challenging, largely due to the lack of principled proof planning and the resulting combinatorial explosion in proof search.
Combinatorics is widely considered one of the most difficult areas for ATP~\cite{chen2025seedproverdeepbroadreasoning,deepmind-gemini-deepthink-imo2025}, and \emph{combinatorial identities} form a fundamental and ubiquitous class of statements within it.
Consequently, automating machine-checkable proofs of combinatorial identities is a key objective for scalable theorem proving, enabling the construction of reusable verified libraries and reducing the human effort required for formal proof development.

We focus on automating formal proofs of \emph{combinatorial identities} in Lean~4~\cite{moura2021lean}, a task that remains a challenge for current LLM-based provers.
Such proofs typically require \textbf{long-horizon proof planning}: without an executable global route that specifies key intermediate milestones, an LLM-based prover is forced into largely unconstrained exploration, leading to a combinatorial explosion in proof search.
Moreover, \textbf{data scarcity} limits the available proof data for this domain.
These challenges motivate an approach for automatically generating \emph{executable proof sketches} that decompose a target identity into a sequence of machine-checkable subgoals, thereby constraining the search space and enabling long-horizon reasoning.

The WZ method is a classical and powerful approach to prove combinatorial identities~\cite{zeilberger1991method,1990WZHerbert,Gosper1978,liu2024deepseek}.
It provides a natural source of \emph{global proof structure}: a WZ proof is organized around the synthesis of a recurrence together with appropriate boundary and initial conditions, thereby providing a principled \emph{proof plan} for long-horizon reasoning~\cite{zeilberger1991method}.
However, turning the WZ method into a fully automated formal proof pipeline remains challenging in practice, due to limited symbolic coverage and the nontrivial formalization of boundary conditions and obligations in proof assistants.
This motivates a sketch-guided neuro-symbolic design: we use the WZ method to \emph{plan and decompose} a target identity into verifiable intermediate goals and leverage LLMs to \emph{discharge} the resulting formal obligations.
Meanwhile, LLM-based proving also extends coverage by producing direct Lean proofs when the WZ decomposition is unavailable or fails.

We propose \textbf{WZ-LLM}, a framework that translates the WZ proof plan into \emph{executable proof sketches} in Lean—namely, recurrence lemmas and their associated obligations—and leverages a large language model to automatically discharge the resulting subgoals.
To make this approach effective in practice and mitigate data scarcity for Lean combinatorics, we develop a Lean-verified training pipeline for WZ-LLM.
We start by manually formalizing \textbf{307} combinatorial identities from classical textbooks, yielding a high-quality seed corpus for cold-start supervised fine-tuning (SFT).
Building on this seed, we expand the data via expert-verified iteration, retaining only \emph{Lean-kernel-verified} model outputs: verified proofs (including WZ-sketch lemmas and direct proofs of WZ-uncovered identities) are added to the training corpus and pruned from the proving-task pool, while unverified attempts are retained for subsequent iterations.
We then train a specialized prover, \textbf{WZ-Prover}, through cold-start SFT, iterative training on expanded data~\cite{InternLM2.5-StepProver}, and Dynamic Sampling Policy Optimization (DAPO) refinement~\cite{yu2025dapo}. 
Experiments show that WZ-LLM achieves a \textbf{34\%} end-to-end success rate on a benchmark of \textbf{100} classical combinatorial identities in Lean~4, substantially outperforming strong baselines.
Furthermore, on \emph{LCI-Test}, WZ-LLM proves \textbf{5} identities on which the symbolic-only baseline fails, highlighting complementary coverage beyond purely symbolic methods.
WZ-LLM also improves performance on CombiBench and PutnamBench, solving all manually identified identity instances in these benchmarks. Our main contributions are as follows.
\begin{itemize}
  \item We propose WZ-LLM, a neuro-symbolic framework that converts Wilf–Zeilberger reasoning into \emph{executable proof sketches} in Lean and uses an LLM-based prover to discharge the resulting proof obligations.
\item We build a Lean-verified dataset for combinatorial identities and train a domain-specialized prover via a multi-stage pipeline (cold-start SFT, verified bootstrapping, and DAPO refinement), improving long-horizon lemma chaining under distribution shift.
  \item We evaluate WZ-LLM on a new benchmark \emph{LCI-Test} (100 classical identities) as well as public combinatorics benchmarks, outperforming strong LLM baselines and complementing a symbolic-only baseline by solving additional cases.
\end{itemize}

\section{Related Work}
 
{\bf Automated Theorem Proving.} LLM-based automated theorem proving has advanced rapidly in interactive proof assistants such as Lean~\cite{moura2021lean} and Isabelle~\cite{paulson1994isabelle}.
One line focuses on step-wise tactic generation with explicit search (e.g., BFS/MCTS) and frequent verifier interaction, achieving strong performance but often incurring substantial branching and compute costs~\cite{polu2020,Leandojo,xin2025bfs,InternLM2.5-StepProver}.
Another line emphasizes whole-proof generation to preserve long-range coherence, but may underutilize intermediate verifier signals for frameworkatic correction~\cite{wang2024theoremllama,lin2025goedel}.
To mitigate data scarcity and improve long-horizon reasoning, recent methods leverage verifier-filtered bootstrapping and reinforcement learning in verifiable environments~\cite{shao2024deepseekmath,ospanov2025apollo,xin2024deepseek}.
In particular, MA-LoT~\cite{wang2025ma} GAR~\cite{wang2025gar} and Spark-Prover-X1~\cite{zhou2025spark} further improve proving via correction/refinement and scalable verifier-guided training.
Our work targets a domain where successful proving critically depends on \emph{explicit proof planning} and studies how symbolic decompositions can be compiled into executable proof sketches to constrain proof search.

{\bf Formal Proof of Combinatorial Identities.} 
 Symbolic computation has long played a central role in proving combinatorial identities by transforming summations into algebraic forms that are amenable to frameworkatic verification.
Classical \emph{generating-function} techniques reduce identities to functional equations and coefficient extraction, and remain a standard paradigm in combinatorics~\cite{stanley1986enumerative,Flajolet_Sedgewick_2009,wilf2005generatingfunctionology}.
Beyond generating functions, elimination-style approaches (e.g., Gröbner-basis methods) establish polynomial or rational identities via ideal membership and symbolic elimination~\cite{DBLP:books/daglib/0091062,sturmfels2002solving}.
 For hypergeometric and holonomic summations, certificate-based methods such as the Wilf--Zeilberger  algorithm and creative telescoping derive recurrences together with rational certificates/invariants whose validity can be mechanically checked~\cite{wilf1992algorithmic, petkovsek1996b,DBLP:series/tmsc/KauersP11}.
However, while these techniques are highly effective in computer algebra frameworks, their outputs do not directly translate into interactive proof assistants:
end-to-end formalization typically requires explicitly reconstructing telescoping arguments, boundary conditions, normalization steps, and non-vanishing side conditions inside the logic, which often dominates the verification cost. Harrison~\cite{Harrison2015} calculated rational function certificates using the Maxima external computer algebra framework, and subsequently used HOL Light \cite{harrison1996hol} to rigorously verify that these certificates satisfy the WZ equation (\ref{eq:wz_pair}). 
Recent efforts have begun to framework this domain in Lean by constructing dedicated datasets and benchmarks for combinatorial identities~\cite{xiong2025combinatorialidentitiesbenchmarktheorem}. In this work, we leverages the WZ method as a decomposition tool that compiles identities into explicit proof obligations, with all proofs verified by Lean~4.

 \begin{figure*}[htp!]
    \centering
    \includegraphics[width=\textwidth]{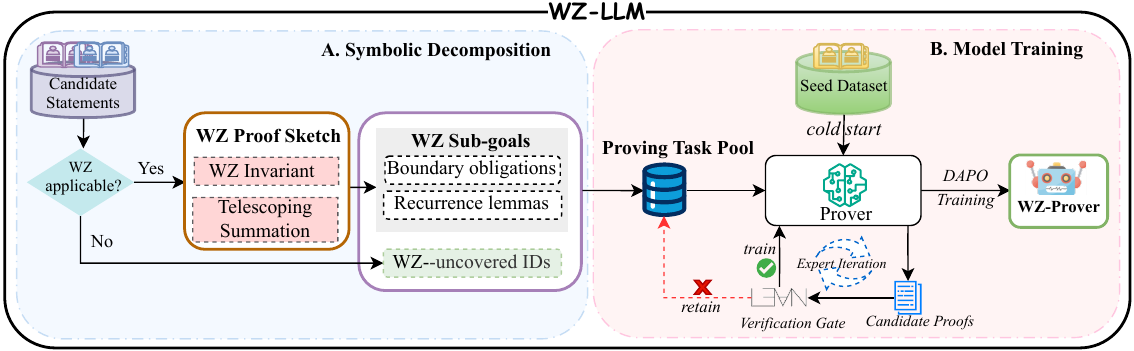} 
    \caption{\textbf{Overview of \textbf{WZ-LLM}.}
\textbf{(A) Symbolic decomposition.} Given combinatorial-identity statements, the symbolic engine checks WZ applicability and, when applicable, constructs a WZ proof sketch that is decomposed into sub-goals. Statements that are not amenable to the WZ method are labeled as \emph{WZ-uncovered} targets. All sub-goals and uncovered targets form a shared pool of proving tasks.
\textbf{(B) Model training.} We cold-start a prover with a seed dataset, iteratively generate proofs for tasks in the pool, and verify them. Lean-verified proofs are retained and used for prune/train updates, while unverified ones are discarded. The expert model is further refined with DAPO to yield \textbf{WZ-Prover} for formally discharging both WZ-derived obligations and WZ-uncovered identities.
}
    \label{fig:wzllm_overview}
\end{figure*}

\section{Preliminaries}
 \label{sec:wz-llm}

 We briefly review the Wilf--Zeilberger  method, a certificate-based symbolic framework to prove or discover definite summation identities~\cite{1990WZHerbert}. Let $F(n,k)$ be a bivariate hypergeometric term, i.e., $F(n+1,k)/F(n,k)$ and $F(n,k+1)/F(n,k)$ are both rational functions of $n$ and $k$. Assume that for each fixed $n$ as a nonnegative integer, $F(n,k)$ vanishes for all but finitely many values of $k$.
The goal of the WZ method is to establish identities of the form
\begin{equation}
\label{eq:wz_target}
S(n) := \sum_{k} F(n,k) = C,
\end{equation}
where the index $k$ runs on all integers.
Note that, for summations of the form \(\sum_k f(n,k) = r(n)\) with \(r(n) \neq 0\), one can normalize the identity by defining \(F(n,k) = f(n,k)/r(n)\), thereby reducing it to the standard form \(\sum_k F(n,k) = 1\). This normalization step is crucial in practice, as it simplifies the proof to a uniform goal of showing that the sum equals a constant and thus facilitates the frameworkatic construction of certificates.

The core of the WZ method is to build a WZ-pair. Specifically, the algorithm seeks an auxiliary term \(G(n,k)\) such that the following WZ equation holds:
\begin{equation}
\label{eq:wz_pair}
F(n+1, k) - F(n, k) = G(n, k+1) - G(n, k).
\end{equation}
Under the appropriate boundary conditions, summing both sides of the equation over \(k\) makes the right-hand side telescope zero. Hence, we have
\[
S(n+1) - S(n) = \sum_{k=-L}^{k} \left[ F(n+1, k) - F(n, k) \right] = 0.
\]
This implies that \(S(n)\) is independent of \(n\), so the identity \eqref{eq:wz_target} remains and the constant C can be determined from an initial value \(S(n_0)\).
It is required that \(G(n,k)\) take the multiplicative form \(G(n,k) = R(n,k)F(n,k)\), where \(R(n,k)\) is a rational function, known as the ``WZ certificate". Such certificates can be found automatically via Gosper' algorithm or Zeilberger creative telescoping.

\begin{tcolorbox}[
  colback=gray!3,
  colframe=black!40,
  boxrule=0.5pt,
  arc=2pt,
  left=6pt,right=6pt,top=6pt,bottom=6pt,
]
Consider the classical binomial-sum identity
\begin{equation}
\label{eq:wz_box_example_target}
\sum_{k=0}^{n} \binom{n}{k}^{2} \;=\; \binom{2n}{n}.
\end{equation}

\smallskip
\noindent\textbf{Step 1 (Normalization).}
Divide both sides by the right-hand side and reduce to the standard form
\begin{equation}
\label{eq:wz_box_example_norm}
\sum_{k=0}^{n} F(n,k) = 1,
\qquad
F(n,k) := \frac{\binom{n}{k}^{2}}{\binom{2n}{n}}.
\end{equation}

\smallskip
\noindent\textbf{Step 2 (Certificate synthesis).}
A symbolic engine synthesizes a rational certificate
\begin{equation}
\label{eq:wz_box_example_cert}
R(n,k)
\;=\;
-\frac{k^{2}\,(3n-2k+3)}{2\,(n-k+1)^{2}\,(2n+1)},
\end{equation}
and defines $G(n,k) := R(n,k)\,F(n,k)$ such that the WZ relation in Eq.(\ref{eq:wz_pair}) holds.

\smallskip
\noindent\textbf{Step 3 (Lean-checkable obligations).}
The proof of the identity is reduced to a small set of structured obligations:
(i) algebraically verify Eq.(\ref{eq:wz_pair});
(ii) show that the boundary terms vanish, i.e., $G(n,0)=G(n,n\!+\!1)=0$;
and (iii) check the base case to conclude that the constant equals 1. 
Once (i)--(iii) hold, summing Eq.~\eqref{eq:wz_pair} over $k$ yields $S(n\!+\!1)-S(n)=0$ and thus Eq.~\eqref{eq:wz_box_example_target} follows.
\end{tcolorbox}

The boxed derivation exposes an \emph{executable proof sketch}: a recurrence/telescoping relation plus
\emph{boundary obligations}, which are exactly the two main types of formal proof task used in our framework.
In this work, we use the WZ method as a principled symbolic certification tool to automatically generate executable proof sketches for combinatorial identities. The resulting sketch transforms the original identity into a set of structured proof obligations, primarily consisting of boundary-condition verification and recurrence-based lemmas. This provides crucial, verifiable structural guidance that constrains the subsequent LLM-based formal proving process and enables scalable proof search in Lean~4.

\section{Methodology}

In this section, we present \textbf{WZ-LLM}, a neuro-symbolic framework for formally proving combinatorial identities in Lean~4.
As shown in Figure~\ref{fig:wzllm_overview}, for identities that are amenable to the WZ method, WZ-LLM first constructs an executable \textbf{WZ proof sketch} and compiles the original goal into a set of structured proving tasks, including boundary-condition obligations and recurrence-based lemmas.
Identities beyond the coverage of the WZ are also incorporated into the task set as direct proving targets.
All tasks are then discharged by our specialized prover model, \textbf{WZ-Prover}, and the resulting proofs are strictly verified by the Lean kernel.
To improve robustness on sketch-induced subgoals and generalization to out-of-distribution identities, we build a broad-coverage formal training corpus with difficulty stratification, and train WZ-Prover via a combination of expert-in-the-loop iterative data expansion, and DAPO refinement with difficulty-smoothing.
In general, \textbf{WZ-LLM} consists of two stages: \emph{(i)} \textit{Symbolic Decomposition} using the WZ method (Section~\ref{sec:4.1}) and \emph{(ii)} \textit{Model Training} for WZ-Prover (Section~\ref{sec:4.2}).

\subsection{Symbolic Decomposition}
\label{sec:4.1}

Given a Lean~4 statement of a combinatorial identity, WZ-LLM constructs an WZ-style proof sketch and compiles it into a finite set of Lean proof obligations (Figure~\ref{fig:wzllm_overview}.A).
Crucially, it not only produces a WZ certificate/recurrence but also generates the auxiliary side conditions and case splits needed for \emph{successful formalization} under Lean's semantics.

\noindent \textbf{Normalization.}
The input is modeled into a canonical summation form (e.g. \texttt{Finset.range (n+1)} with index starting at \(0\)) using standard transformations:
(i) shifting summation ranges not starting at \(0\);
(ii) rewriting bounded sums (\texttt{Icc}, \texttt{Ico}) into \texttt{range-style} sums when possible; (iii) algebraic normalization of hypergeometric factors (factorials/binomials/powers) to reduce syntactic variance.
For piecewise specifications (e.g., parity constraints), we compile the goal into structured case splits (e.g. \(n\) even/odd) and treat each branch as an independent proving task.

\noindent \textbf{WZ proof sketch construction.}
When the goal is covered by the WZ method, a symbolic engine proposes a recurrence/certificate (e.g., a rational function \(R(n,k)\) with \(G(n,k)=R(n,k)F(n,k)\)).
The sketcher then instantiates the telescoping relation and reduces the original identity to standard WZ obligations:
(i) a recurrence lemma showing that the normalized sum \(S(n)\) is invariant (or satisfies a linear recurrence) 
and (ii) boundary obligations ensuring that the telescoping sum collapses to boundary terms.

\noindent \textbf{Constraint inference and side-condition generation.}
A major practical obstacle in Lean formalization is that algebraic steps (e.g., \texttt{field\_simp}) require explicit non-vanishing assumptions for denominators and normalization factors.
We therefore automatically infer \emph{domain constraints} for each obligation, including when summands or normalizers become zero/undefined, and compile them into Lean side-condition goals.
Concretely, we use symbolic simplification and constraint solving to identify problematic points (e.g. zero denominators at boundary indices, negative arguments to factorial-like terms, or sign-sensitive rewrites), and then
(i) generate corresponding nonzero lemmas (e.g. \(\forall n,k, A(n,k)\neq 0\), \(\forall n, B(n)\neq 0\)),
(ii) isolate boundary indices as explicit subgoals,
and (iii) trigger case splits when constraints depend on discrete properties such as parity.

\noindent \textbf{Proving Task Pool.}
Overall, each identity is compiled into a proving-task set
\(\mathcal{T}=\mathcal{T}_{\mathrm{rec}}\cup\mathcal{T}_{\mathrm{bd}}\cup\mathcal{T}_{\mathrm{side}}\cup\mathcal{T}_{\mathrm{norm}}\cup\mathcal{T}_{\mathrm{case}}\),
where \(\mathcal{T}_{\mathrm{rec}}\) contains recurrence lemmas, \(\mathcal{T}_{\mathrm{bd}}\) contains boundary obligations,
\(\mathcal{T}_{\mathrm{side}}\) contains non-vanishing and well-typedness conditions required by algebraic tactics,
\(\mathcal{T}_{\mathrm{norm}}\) contains normalization/range-shift lemmas, and \(\mathcal{T}_{\mathrm{case}}\) contains case-split branches.
All obligations are expressed as Lean goals and are sent to \textbf{WZ-Prover} for formal discharge.

\subsection{Model Training Pipeline for \textbf{WZ-Prover}}
\label{sec:4.2}

The symbolic stage in Section~\ref{sec:4.1} compiles a target identity into a finite set of Lean-checkable proving tasks, including boundary-condition obligations, lemmas based on recurrence, and direct identity goals.
If a valid symbolic sketch can be synthesized, releasing these obligations yields an end-to-end Lean proof of the original identity.
However, symbolic certificate synthesis may fail, and many identities fall outside the WZ-applicable class.
To bridge this gap, we train a combinatorics-specialized Lean prover, \textbf{WZ-Prover}, that covers both:
(i) \emph{sub-goals} produced by symbolic decomposition, and
(ii) \emph{WZ-uncovered identities} that must be proved directly without symbolic guidance.
Training proceeds in three stages:
(1) supervised cold-start (SFT) on Expert seed dataset;
(2) expert-in-the-loop bootstrapping that expands the training set with new Lean-verified lemma and identity proofs;
and (3) verifier-based reinforcement learning via DAPO to improve exploration and robustness on hard proving tasks. 

\noindent
\textbf{Data construction.} 
We adopt a data-flow strategy that combines a small high-quality seed dataset, a large pool of unlabeled candidate combinatorial identity statements, and expert-in-the-loop iterative expansion, while retaining a \textbf{source-disjoint} test set to evaluate cross-source generalization. The resulting training and test sets are named \textbf{LCI-training} and \textbf{LCI-Test}, respectively. Specifically, our dataset consists of three components: (i) \textbf{Expert seed dataset.} Based on a classical reference~\cite{spivey2019art}, we manually formalized $307$ combinatorial identities and provided complete Lean proofs, ensuring all samples are verified in Lean without \texttt{sorry}. We further applied a symbolic decomposition method to split these identities into CI proving tasks, generating a total of $1200$ sub-goals; (ii) \textbf{Candidate statements pool.} From two additional classical sources~\cite{CISTIRbook,gould_quaintance_2010_tables}, we extracted and formalized $1020$ identity statements in Lean 4 as candidate training targets; and (iii) \textbf{Source-disjoint test set.} We exclusively reserve two classical combinatorial identity collections~\cite{gould1972combinatorial,shi2001combinatorial} for evaluation, selecting $100$ problems as the test set. To further reduce potential data leakage, expressions are canonized to ensure that the canonical forms of test identities do not appear in the candidate pool or training set.

\subsubsection{Expert-in-the-loop bootstrapping}

The baseline model is cold-started on a high-quality expert seed dataset (307 identities with complete Lean proofs, plus 1200 framework-extracted lemma proofs) to acquire basic proving capability. WZ-LLM then generates proofs for the candidate statement pool from this cold-start model via two routes—symbolic decomposition and direct proving—and we retain only samples that compile and verify in Lean (deduplicated before training).

This process yields two types of new training data: (i) intermediate lemma proofs for WZ sub-goals, and (ii) fully formalized identity proofs. In Round~1, we sample 1020 identities; symbolic decomposition produces 7385 lemma obligations, and we retain 5139 verified lemma proofs plus 32 verified end-to-end proofs. In Round~2, rerunning on the remaining goals yields 532 additional lemma proofs and 79 identity proofs, so expert iteration adds 5671 lemma proofs and 111 identity proofs in total.
Notably, the intermediate lemmas align with key steps in the symbolic trajectory, while the full identity proofs help preserve and improve end-to-end performance.

\subsubsection{DAPO with Difficulty-Smoothing}
\label{app:dapo_refine}

We further refine \textsc{WZ-Prover} with a reinforcement-learning (RL) stage to improve robustness on long-horizon lemma chaining and hard end-to-end identities.
This refinement has two parts: (i) constructing a difficulty-smoothed RL set from the SFT corpus, and (ii) applying Dynamic Action Policy Optimization (DAPO)~\cite{yu2025dapo}, a stable Long-CoT optimizer that mitigates entropy collapse, truncation-induced reward noise, and training instability.

Our supervised corpus combines (a) a high-quality expert seed set and (b) Lean-verified samples from expert-verified iteration over a pool of candidate statements, including both sketch-induced subgoals (symbolic lemmas) and full identity goals.
These goal types have markedly different difficulty profiles: symbolic decomposition yields many short and often repetitive obligations, whereas full identities (and hard subgoals) require longer reasoning chains and fail more frequently.
To avoid RL overfitting to frequent easy patterns, we run a rollout-based diagnostic pass to estimate a \emph{pass rate} for each goal under the current policy, use it to tag goals into coarse difficulty bins, and filter the extremes by downsampling near-duplicate easy instances and removing near-zero-pass-rate instances that contribute mostly noisy gradients.
The resulting RL set preserves diverse mid-to-hard goals while maintaining a smoother difficulty distribution.

We then optimize with DAPO using a rule-based outcome reward from Lean verification together with a soft overlong punishment:
\begin{equation}
\label{eq:dapo_total_reward_final}
R(\pi;G)=R_{\text{out}}(\pi;G)+\lambda_{\text{len}}\,R_{\text{len}}(\pi),
\end{equation}
where $R_{\text{out}}(\pi;G)\in\{+1,-1\}$ indicates whether $\pi$ kernel-verifies $G$, and $R_{\text{len}}$ applies a gradual penalty near the maximum token budget.

\section{Experiments}

  We conduct comprehensive experiments on the classical combinatorial-identity benchmark \textbf{LCI-Test} and two external benchmarks, \textbf{CombiBench} and \textbf{PutnamBench-Comb} (the combinatorics subset of PutnamBench), to evaluate the ability of \textbf{WZ-LLM} to produce end-to-end formal proofs in Lean~4. Our evaluation focuses on whether, under a fixed sampling budget, the framework can leverage \emph{proof-sketch} guidance to generate longer and more structured proof chains, and improve success rates on identities outside the coverage of symbolic computation methods (WZ-uncovered).
Specifically, Section~\ref{exp:main} reports the main results on LCI-Test, establishing the overall effectiveness and generality of \textbf{WZ-LLM}. Section~\ref{exp:comb_put} further evaluates its generalization to broader combinatorics problems on CombiBench and PutnamBench-Comb.    Finally, Section~\ref{exp:train} presents ablations that isolate the contributions of key training components.

\begin{table*}[t]
\centering
\caption{\textbf{Main results on LCI-Test.}
We evaluate our framework \textbf{WZ-LLM}, which combines a symbolic \textbf{WZ-Sketch} decomposer with a specialized Lean prover \textbf{WZ-Prover} trained from Goedel-Prover-V2.
The upper block reports prior baselines under their respective decoding budgets.
The \emph{Ours} block is organized into two parts.
\textbf{\emph{(I) Component effects:}} WZ-Sketch +Goedel-Prover-V2 matches the Goedel-Prover-V2 baseline, while WZ-Prover improves direct proving over the same pass@32 budget.
\textbf{\emph{(II) Two-route solving:}} \textbf{WZ-uncovered} counts identities where WZ decomposition is inapplicable and are solved directly by WZ-Prover; \textbf{WZ-Sketch}+\textbf{WZ-Prover} counts WZ-applicable identities solved via sketch-induced obligations discharged by WZ-Prover.
\textbf{WZ-LLM} aggregates both routes, achieving the best overall success (34/100).} 
\label{tab:main}
\begin{tabular}{l c c c}
\toprule
\textbf{Method}& \textbf{Model size} & \textbf{Sample budget} & \textbf{LCI-Test}   \\
\midrule
Gemini-3.1-Pro-Preview &- & pass@32 &16/100\\
DeepSeek-V3~\cite{liu2024deepseek}&685B & pass@32 &1/100\\
InternLM-2.5-StepProver~\cite{InternLM2.5-StepProver}  &7B & $4 \times 32 \times 600$  & 2/100 \\

MA-LoT ~\cite{wang2025ma}       &7B           & $16 + 8 \times 2$                        & 3/100  \\
Kimina-Prover-Distill ~\cite{wang2025kimina}&7B & pass@32                         & 6/100   \\
DeepSeek-Prover-V2 ~\cite{ren2025deepseek}  &7B   & pass@32                        & 6/100   \\
Goedel-Prover-V2 ~\cite{lin2025goedel}   &8B   & pass@32                        & 9/100   \\
\midrule
\multicolumn{3}{l}{\textit{Ours}}  \\
\midrule  
\textbf{WZ-Sketch + Goedel-Prover-V2}   &8B              & pass@32 & 9/100    \\
\textbf{WZ-Prover}    &8B              & pass@32 & \textbf{12/100} \\
\cdashline{1-4}
\textbf{WZ-uncovered}     &8B              & pass@32 & \textbf{5/100} \\
\textbf{WZ-Sketch} + \textbf{WZ-Prover}   &8B              & pass@32 & \textbf{29/100 }   \\
\cdashline{1-4}
\textbf{WZ-LLM}     &8B              & pass@32 & \textbf{34/100} \\  
\bottomrule
\end{tabular}
\end{table*}

\subsection{Experiment Setup}

\subsubsection{Benchmark and Baselines}
\label{sec:5.1}

We evaluate \textbf{WZ-LLM} on three benchmarks: our in-domain benchmark \textbf{LCI-Test}, and two external benchmarks, \textbf{CombiBench} and \textbf{PutnamBench-Comb}. \textbf{LCI-Test} is a combinatorial-identity benchmark constructed from classical references~\cite{gould1972combinatorial, shi2001combinatorial}, containing 100 identities to assess the ability of WZ-LLM to formally prove combinatorial identities in Lean~4. \textbf{CombiBench} is a widely used and challenging benchmark for formal combinatorics theorem proving, consisting of 100 combinatorial problems~\cite{liu2025combibench}. PutnamBench~\cite{tsoukalas2024putnambench}, which contains 672 Lean~4-formalized Putnam problems, we extract a combinatorics subset of 36 problems annotated with the \texttt{combinatorics} tag, which we refer to as \textbf{PutnamBench-Comb}. These external benchmarks evaluate WZ-LLM's generalization to broader combinatorics problems.

We compare WZ-LLM against a closed-source LLM, DeepSeek-V3~\cite{liu2024deepseek}, and open-source provers, including Kimina-Prover-Distill~\cite{wang2025kimina}, DeepSeek-Prover-V1.5-RL~\cite{xin2024deepseek}, DeepSeek-Prover-V2~\cite{guo2025deepseek}, Goedel-Prover-V2~\cite{lin2025goedel}, MA-LoT~\cite{wang2025ma}, and InternLM-2.5-StepProver~\cite{InternLM2.5-StepProver}.

\subsubsection{Implementation Details}
For symbolic decomposition, we use \textsc{SageMath}~\cite{sage2005} as the underlying computer algebra framework (CAS) to carry out symbolic computations in the WZ/creative-telescoping pipeline. In particular, we rely on the hypergeometric summation routines provided by the \texttt{sage.combinat} module: given a bivariate hypergeometric term $F(n,k)$, the corresponding WZ certificate $C(n,k)$ is synthesized via \texttt{F.WZ\_certificate(n,k)}, and the recurrence ratios are simplified using \texttt{simplify()}.

On all benchmarks, we report end-to-end proof success rates under the pass@32 metric: a run is counted as successful only if the \emph{entire} proof is accepted by the Lean~4 kernel.
The total training cost is 16 GPU-days, and the evaluation cost is 9 GPU-days, all conducted on a 4$\times$ L40s-48GB GPU cluster.
To quantify the contribution of each component, we report two ablation variants:
(i) \textbf{WZ-Sketch}, which retains WZ proof-sketch guidance while removing our specialized training pipeline; and
(ii) \textbf{WZ-Prover}, which retains our training pipeline and specialized prover while disabling WZ proof-sketch guidance.
These ablations isolate the gains from symbolic sketching and prover specialization, respectively.
In the full \textsc{WZ-LLM} framework, goals are solved via two complementary routes. The \emph{sketch-guided} route invokes symbolic WZ decomposition to compile an identity into a set of Lean-checkable lemma obligations, which are then discharged one by one by \textbf{WZ-Prover} (reported as \textbf{WZ-Sketch + WZ-Prover} in the table). The \emph{direct} route, used when decomposition is inapplicable or fails, attempts an end-to-end Lean proof directly with \textbf{WZ-Prover} (reported as \textbf{WZ-uncovered}). Finally, we report \textbf{WZ-LLM} as the union of successes from these two routes.

\subsection{Main Results}
\label{exp:main}

Table~\ref{tab:main} reports end-to-end performance on the classic combinatorial identity benchmark LCI-Test (100 problems), measured by the number of fully verified Lean4 proofs. Overall, LCI-Test remains challenging for existing provers under comparable sampling budgets: DeepSeek-V3 solves 1/100, InternLM-2.5-StepProver reaches 2/100, and among whole-proof provers evaluated with pass@32, Kimina-Prover-Distill and DeepSeek-Prover-V2 both solve 6/100. Goedel-Prover-V2 provides the strongest pass@32 baseline at 9/100.

\noindent
\textbf{Component effects.}
Building on Goedel-Prover-V2, we analyze WZ-LLM through two complementary views. 
 Adding \textbf{WZ-Sketch} on top of the Goedel-Prover-V2 backend does not improve end-to-end results (9/100), indicating that sketch-only decomposition is insufficient without a stronger lemma prover. In contrast, our trained \textbf{WZ-Prover} improves \emph{direct} end-to-end proving to 12/100 under the same pass@32 budget, demonstrating that the proposed specialization pipeline strengthens the model’s ability to discharge core algebraic subgoals even without symbolic guidance.

\noindent
\textbf{Two-route solving.} Our full framework WZ-LLM combines (i) a sketch-guided route for WZ-applicable identities, where \textbf{WZ-Sketch} decomposes the goal into Lean-checkable obligations that are discharged by \textbf{WZ-Prover}, and (ii) a direct route for \textbf{WZ-uncovered} identities, where WZ decomposition is inapplicable and \textbf{WZ-Prover} attempts full proofs end-to-end. On LCI-Test, the WZ-uncovered route solves 5 problems, while the sketch-guided route solves 29 problems, yielding 34 solved problems in total when aggregated in \textbf{WZ-LLM}.

This contrast indicates that our framework provides two complementary capabilities: on the one hand, the specialized \textbf{WZ-Prover} substantially strengthens direct proving on problems \emph{outside the scope of WZ}; on the other hand, \textbf{WZ}-style symbolic decomposition compiles complex identities into a set of Lean-checkable subgoals, and—when paired with a strong lemma prover—effectively converts the benefits of symbolic computation into markedly higher end-to-end proof success. By integrating these two routes within a single framework, \textbf{WZ-LLM} achieves a strong overall end-to-end success rate of $34\%$ on the combinatorial identity benchmark.

\begin{table}[htp!]
\centering
\caption{Lemma-proving performance on Symbolic Decomposition subproblems about the base model with our WZ-Prover. WZ-Prover achieves substantially higher lemma accuracy and solves more end-to-end problems.}
\label{tab:lemma_diag}
\begin{tabular}{l c c c}
\toprule
\textbf{Model} &
\textbf{\#Proved} &
\textbf{Acc.} &
\textbf{\#Solved} \\
\midrule
Goedel-Prover-V2     & 564 / 1{,}178 & 47.88\% & 0 / 46 \\
WZ-Prover  & 864 / 1{,}178 & 73.34\% & 29 / 46 \\
\bottomrule
\end{tabular} 
\end{table}

To understand why sketch-only guidance fails with the untrained backend, we further perform lemma-level diagnostics on the same 1{,}178 sketch-induced subgoals (Table~\ref{tab:lemma_diag}). Goedel-Prover-V2 proves only 564 lemmas (47.88\%), which is insufficient to complete any of the 46 sketched problems because all required obligations must be discharged. Replacing it with \textbf{WZ-Prover} raises lemma coverage to 73.34\% (864/1{,}178), enabling symbolic decomposition to translate into 29 end-to-end solutions. Overall, these results show that WZ-style decomposition is most effective when paired with a specialized lemma prover, and that combining sketch-guided and direct proving routes yields the strongest performance.

\begin{table*}[htp!]
\centering
\caption{\textbf{Cross-dataset evaluation (pass@32) on CombiBench and PutnamBench-Comb.}
All our variants are built on Goedel-Prover-V2 as the base model. The table separates (i) \emph{component ablations} (\textit{WZ-Sketch+Goedel} vs.\ \textit{WZ-Prover}) and (ii) \emph{two-route accounting} in \textbf{WZ-LLM}: the \textbf{WZ-uncovered} route solves goals where WZ decomposition is inapplicable, while the \textbf{WZ-Sketch+WZ-Prover} route solves WZ-applicable goals via symbolic decomposition plus lemma discharge. \textbf{WZ-LLM} aggregates both routes.}
\label{tab:combi_putnam}
\begin{tabular}{lcccc}
\toprule
\textbf{Method} & \textbf{Model size} & \textbf{Sample budget} & \textbf{CombiBench} & \textbf{PutnamBench-Comb} \\
\midrule
Kimina-Prover Distill-8B  & 8B & pass@32 & 6/100  & 0/36 \\
DeepSeek-Prover V2-7B    & 7B & pass@32 & 8/100  & 2/36 \\
Goedel-Prover-V2-8B      & 8B & pass@32 & 12/100 & 0/36 \\
\midrule
\multicolumn{5}{l}{\textit{Ours}} \\
\midrule
\textbf{WZ-Sketch + Goedel-Prover-V2}  & 8B  & pass@32 & 13/100 & 0/36 \\
\textbf{WZ-Prover} & 8B  & pass@32 & 15/100 & 3/36 \\
\cdashline{1-5} 
\textbf{WZ-uncovered}     &8B              & pass@32 & 15/100 & 2/36 \\
\textbf{WZ-Sketch} + \textbf{WZ-Prover}   &8B              & pass@32 & 1/100  & 1/36 \\
\cdashline{1-5} 
\textbf{WZ-LLM} & 8B  & pass@32 & \textbf{16/100} & \textbf{3/36} \\
\bottomrule
\end{tabular}
\end{table*}

\subsection{Cross-Dataset Evaluation}
\label{exp:comb_put}
Table~\ref{tab:combi_putnam} evaluates cross-dataset generalization on CombiBench and PutnamBench-Comb under a fixed pass@32 sampling budget. Baseline provers remain limited on these heterogeneous combinatorial benchmarks: on CombiBench, Kimina-Prover Distill-8B, DeepSeek-Prover-V2, and Goedel-Prover-V2 solve 6/100, 8/100, and 12/100 problems, respectively; on PutnamBench-Comb, they solve 0/36, 2/36, and 0/36 problems.

\noindent
\textbf{Component effects.}
Using \textbf{WZ-Sketch} with the baseline Goedel-Prover-V2 provides little benefit: Goedel-Prover-V2 alone solves 12/100 problems on CombiBench, and the sketch-guided variant adds only one extra proof (13/100), while on PutnamBench-Comb it offers no improvement (remaining at 0/36). This indicates that symbolic decomposition alone does not translate into end-to-end success when the backend cannot reliably discharge the induced lemma obligations. In contrast, our trained \textbf{WZ-Prover} substantially improves \emph{direct} proving, reaching 15/100 on CombiBench and 3/36 on PutnamBench-Comb, outperforming Goedel-Prover-V2 by +3 and +3 problems, respectively.

\noindent
\textbf{Two-route accounting in \textbf{WZ-LLM}.}
We decompose \textbf{WZ-LLM}'s performance into its two proof routes. On CombiBench, the \textbf{WZ-uncovered} (WZ-inapplicable) route contributes 15 solved problems via direct proving, while the sketch-guided route (\textbf{WZ-Sketch+WZ-Prover}) contributes 1 additional solved problem through symbolic decomposition and lemma discharge, yielding \textbf{16/100} overall. On PutnamBench-Comb, direct proving solves 2/36 problems, and the sketch-guided route adds 1/36, resulting in \textbf{3/36}. 

Overall, these results reinforce the same conclusion as on LCI-Test: \textbf{WZ-Prover} is the main driver of generalization via stronger direct proof capability, while sketch-guided decomposition can provide additive gains when WZ decomposition is applicable and the resulting obligations are within the lemma prover's reach.
\begin{table}[htp!]
\centering
\small
\setlength{\tabcolsep}{6pt}
\begin{tabular}{lccc}
\toprule
\multirow{2}{*}{\textbf{Training stage }} &
\multicolumn{3}{c}{\textbf{LCI-Test }} \\
\cmidrule(lr){2-4}
& pass@1 & pass@8 & pass@32 \\
\midrule
\textit{ SFT (seed only) }                & 1/100 & 3/100 & 9/100  \\
\quad + \textit{expert-iteration}    & 3/100 & 6/100 & 10/100 \\
\quad + \textit{DAPO refinement}                  & 4/100 & 6/100 & 12/100 \\
\bottomrule
\end{tabular}
\caption{\textbf{Training-stage ablations for WZ-Prover on LCI-Test.} }
\label{tab:train_ablation}
\end{table}

\subsection{Effect of Training Stages}
\label{exp:train}

Table~\ref{tab:train_ablation} reports stage-wise ablations for \textsc{WZ-Prover} on LCI-Test, isolating the effects of seed SFT, expert-verified iteration, and DAPO refinement. The cold-start model trained on the seed set alone achieves only 1/100 at pass@1, 3/100 at pass@8, and 9/100 at pass@32, indicating that the expert seed supervision is insufficient to generalize to the harder identities in LCI-Test. Incorporating expert-verified iteration improves performance to 3/100, 6/100, and 10/100, consistent with verifier-filtered data growth that introduces additional WZ-induced obligations and diverse identity proofs (see Appendix~\ref{app:exp_details}). After deduplication and filtering, the resulting SFT corpus contains 418 Lean-verified identities and roughly 5{,}000 lemma proofs (about 5{,}418 samples in total).

Finally, DAPO refinement further raises performance to 4/100, 6/100, and 12/100. Notably, the improvement is concentrated at larger sampling budgets (pass@32), which suggests that RL primarily helps on harder instances that require longer proof trajectories and benefit from increased exploration, rather than on the easiest problems already solved by SFT. Additional implementation details for the RL set construction (difficulty smoothing via pass-rate filtering) and DAPO settings are provided in Appendix~\ref{app:exp_details}.

\section{Conclusion}
We propose WZ-LLM, a neuro-symbolic framework that integrates the Wilf--Zeilberger method with LLM-based proving for combinatorial identities in Lean~4.
WZ-LLM translates  WZ proof plans into {executable proof sketches} 
and uses a trained prover, WZ-Prover, to discharge the resulting machine-checkable subgoals.  This approach provides principled long-horizon proof planning while extending coverage beyond purely symbolic methods. Across multiple benchmarks, WZ-LLM achieves higher proof success rates than strong LLM baselines; on LCI-Test, it additionally proves identities that defeat a symbolic-only baseline. Overall, these results indicate that coupling symbolic proof planning with learned formal reasoning is a promising direction for machine-checkable scalable verification in combinatorics.

\section*{Impact Statement}

This paper presents a Lean-verified proving framework that combines symbolic WZ decomposition with a trained neural prover. The approach can improve automation for formalizing combinatorial identities and reduce the manual effort needed to obtain kernel-checked proofs. We do not anticipate direct negative societal impacts beyond standard concerns about overreliance on automated tools; all results are validated by the Lean 4 kernel.

\nocite{langley00}
\bibliographystyle{icml2026}
\bibliography{WZ-LLM}

\begin{thebibliography}{46}
\providecommand{\natexlab}[1]{#1}
\providecommand{\url}[1]{\texttt{#1}}
\expandafter\ifx\csname urlstyle\endcsname\relax
  \providecommand{\doi}[1]{doi: #1}\else
  \providecommand{\doi}{doi: \begingroup \urlstyle{rm}\Url}\fi

\bibitem[Chen et~al.(2025)Chen, Gu, Huang, Huang, Jiang, Jie, Jin, Jin, Li, Ma, Ren, Shen, Shi, Sun, Sun, Wang, Wang, Wang, Wei, Wei, Wu, Wu, Xia, Xin, Yang, Ying, Yuan, Yuan, Zhan, Zhang, Zhang, Zhang, Zhao, Zhao, Zhou, and Zhu]{chen2025seedproverdeepbroadreasoning}
Chen, L., Gu, J., Huang, L., Huang, W., Jiang, Z., Jie, A., Jin, X., Jin, X., Li, C., Ma, K., Ren, C., Shen, J., Shi, W., Sun, T., Sun, H., Wang, J., Wang, S., Wang, Z., Wei, C., Wei, S., Wu, Y., Wu, Y., Xia, Y., Xin, H., Yang, F., Ying, H., Yuan, H., Yuan, Z., Zhan, T., Zhang, C., Zhang, Y., Zhang, G., Zhao, T., Zhao, J., Zhou, Y., and Zhu, T.~H.
\newblock Seed-prover: Deep and broad reasoning for automated theorem proving.
\newblock 2025.
\newblock URL \url{https://arxiv.org/abs/2507.23726}.

\bibitem[Cox et~al.(1997)Cox, Little, and O'Shea]{DBLP:books/daglib/0091062}
Cox, D.~A., Little, J., and O'Shea, D.
\newblock \emph{Ideals, varieties, and algorithms - an introduction to computational algebraic geometry and commutative algebra {(2.} ed.)}.
\newblock Undergraduate texts in mathematics. Springer, 1997.
\newblock ISBN 978-0-387-94680-1.

\bibitem[Flajolet \& Sedgewick(2009)Flajolet and Sedgewick]{Flajolet_Sedgewick_2009}
Flajolet, P. and Sedgewick, R.
\newblock \emph{Analytic Combinatorics}.
\newblock Cambridge University Press, 2009.

\bibitem[{Google DeepMind}(2025)]{deepmind-gemini-deepthink-imo2025}
{Google DeepMind}.
\newblock Advanced version of gemini with deep think officially achieves gold medal standard at the international mathematical olympiad.
\newblock Technical report, Google DeepMind, 2025.
\newblock URL \url{https://deepmind.google/blog/advanced-version-of-gemini-with-deep-think-officially-achieves-gold-medal-standard-at-the-international-mathematical-olympiad/}.
\newblock Accessed: 2026-01-15.

\bibitem[Gosper(1978)]{Gosper1978}
Gosper, R.
\newblock Decision procedure for indefinite hypergeometric summation.
\newblock \emph{Proceedings of the National Academy of Sciences of the United States of America}, 75\penalty0 (1):\penalty0 40—42, January 1978.
\newblock ISSN 0027-8424.
\newblock \doi{10.1073/pnas.75.1.40}.
\newblock URL \url{https://europepmc.org/articles/PMC411178}.

\bibitem[Gould(1972)]{gould1972combinatorial}
Gould, H.~W.
\newblock \emph{Combinatorial Identities}.
\newblock West Virginia University, Morgantown, WV, 1972.

\bibitem[Gould(2010)]{gould_quaintance_2010_tables}
Gould, H.~W.
\newblock Tables of combinatorial identities (from the seven unpublished manuscripts of h. w. gould).
\newblock Unpublished manuscript collection, 2010.
\newblock URL \url{https://math.wvu.edu/~hgould/}.
\newblock Edited and compiled by Jocelyn Quaintance (May 3, 2010). Collection of PDF volumes hosted online.

\bibitem[Guo et~al.(2025)Guo, Yang, Zhang, Song, Zhang, Xu, Zhu, Ma, Wang, Bi, et~al.]{guo2025deepseek}
Guo, D., Yang, D., Zhang, H., Song, J., Zhang, R., Xu, R., Zhu, Q., Ma, S., Wang, P., Bi, X., et~al.
\newblock Deepseek-r1: Incentivizing reasoning capability in llms via reinforcement learning.
\newblock \emph{arXiv preprint arXiv:2501.12948}, 2025.

\bibitem[Harrison(1996)]{harrison1996hol}
Harrison, J.
\newblock Hol light: A tutorial introduction.
\newblock In \emph{International Conference on Formal Methods in Computer-Aided Design}, pp.\  265--269. Springer, 1996.

\bibitem[Harrison(2015)]{Harrison2015}
Harrison, J.
\newblock Formal proofs of hypergeometric sums.
\newblock \emph{Journal of Automated Reasoning}, 55\penalty0 (3):\penalty0 223--243, 2015.
\newblock \doi{10.1007/s10817-015-9338-0}.

\bibitem[Kauers \& Paule(2011)Kauers and Paule]{DBLP:series/tmsc/KauersP11}
Kauers, M. and Paule, P.
\newblock \emph{The Concrete Tetrahedron - Symbolic Sums, Recurrence Equations, Generating Functions, Asymptotic Estimates}.
\newblock Texts {\&} Monographs in Symbolic Computation. Springer, 2011.
\newblock ISBN 978-3-7091-0444-6.
\newblock \doi{10.1007/978-3-7091-0445-3}.
\newblock URL \url{https://doi.org/10.1007/978-3-7091-0445-3}.

\bibitem[Langley(2000)]{langley00}
Langley, P.
\newblock Crafting papers on machine learning.
\newblock In Langley, P. (ed.), \emph{Proceedings of the 17th International Conference on Machine Learning (ICML 2000)}, pp.\  1207--1216, Stanford, CA, 2000. Morgan Kaufmann.

\bibitem[Lin et~al.(2025)Lin, Tang, Lyu, Yang, Chung, Zhao, Jiang, Geng, Ge, Sun, et~al.]{lin2025goedel}
Lin, Y., Tang, S., Lyu, B., Yang, Z., Chung, J.-H., Zhao, H., Jiang, L., Geng, Y., Ge, J., Sun, J., et~al.
\newblock Goedel-prover-v2: Scaling formal theorem proving with scaffolded data synthesis and self-correction.
\newblock \emph{arXiv preprint arXiv:2508.03613}, 2025.

\bibitem[Liu et~al.(2024)Liu, Feng, Xue, Wang, Wu, Lu, Zhao, Deng, Zhang, Ruan, et~al.]{liu2024deepseek}
Liu, A., Feng, B., Xue, B., Wang, B., Wu, B., Lu, C., Zhao, C., Deng, C., Zhang, C., Ruan, C., et~al.
\newblock Deepseek-v3 technical report.
\newblock \emph{arXiv preprint arXiv:2412.19437}, 2024.

\bibitem[Liu et~al.(2025)Liu, Lin, Bayer, Dillies, Jiang, Liang, Soletskyi, Wang, Xie, Xiong, et~al.]{liu2025combibench}
Liu, J., Lin, X., Bayer, J., Dillies, Y., Jiang, W., Liang, X., Soletskyi, R., Wang, H., Xie, Y., Xiong, B., et~al.
\newblock Combibench: Benchmarking llm capability for combinatorial mathematics.
\newblock \emph{arXiv preprint arXiv:2505.03171}, 2025.

\bibitem[Moura \& Ullrich(2021)Moura and Ullrich]{moura2021lean}
Moura, L.~d. and Ullrich, S.
\newblock The lean 4 theorem prover and programming language.
\newblock In \emph{Automated Deduction--CADE 28: 28th International Conference on Automated Deduction, Virtual Event, July 12--15, 2021, Proceedings 28}, pp.\  625--635. Springer, 2021.

\bibitem[Ospanov et~al.(2025)Ospanov, Farnia, and Yousefzadeh]{ospanov2025apollo}
Ospanov, A., Farnia, F., and Yousefzadeh, R.
\newblock Apollo: Automated llm and lean collaboration for advanced formal reasoning.
\newblock \emph{arXiv preprint arXiv:2505.05758}, 2025.

\bibitem[Paulson(1994)]{paulson1994isabelle}
Paulson, L.~C.
\newblock \emph{Isabelle: A generic theorem prover}.
\newblock Springer, 1994.

\bibitem[Petkovsek et~al.(1996)Petkovsek, Wilf, and Zeilberger]{petkovsek1996b}
Petkovsek, M., Wilf, H., and Zeilberger, D.
\newblock \emph{A = B}.
\newblock CRC Press, 1996.
\newblock ISBN 9781040187944.
\newblock URL \url{https://books.google.com.nf/books?id=I4qNEQAAQBAJ}.

\bibitem[Polu \& Sutskever(2020)Polu and Sutskever]{polu2020}
Polu, S. and Sutskever, I.
\newblock Generative language modeling for automated theorem proving.
\newblock \emph{arXiv preprint arXiv:2009.03393}, 2020.

\bibitem[Quaintance \& Gould(2015)Quaintance and Gould]{CISTIRbook}
Quaintance, J. and Gould, H.
\newblock \emph{Combinatorial Identities for Stirling Numbers: The Unpublished Notes of H W Gould}.
\newblock 08 2015.
\newblock ISBN 978-981-4725-26-2.
\newblock \doi{10.1142/9821}.

\bibitem[Ren et~al.(2025)Ren, Shao, Song, Xin, Wang, Zhao, Zhang, Fu, Zhu, Yang, et~al.]{ren2025deepseek}
Ren, Z., Shao, Z., Song, J., Xin, H., Wang, H., Zhao, W., Zhang, L., Fu, Z., Zhu, Q., Yang, D., et~al.
\newblock Deepseek-prover-v2: Advancing formal mathematical reasoning via reinforcement learning for subgoal decomposition.
\newblock \emph{arXiv preprint arXiv:2504.21801}, 2025.

\bibitem[Shao et~al.(2024)Shao, Wang, Zhu, Xu, Song, Bi, Zhang, Zhang, Li, Wu, et~al.]{shao2024deepseekmath}
Shao, Z., Wang, P., Zhu, Q., Xu, R., Song, J., Bi, X., Zhang, H., Zhang, M., Li, Y., Wu, Y., et~al.
\newblock Deepseekmath: Pushing the limits of mathematical reasoning in open language models.
\newblock \emph{arXiv preprint arXiv:2402.03300}, 2024.

\bibitem[Shi(2001)]{shi2001combinatorial}
Shi, J.
\newblock \emph{Combinatorial Identities}.
\newblock University of Science and Technology of China Press, 2001.

\bibitem[Spivey(2019)]{spivey2019art}
Spivey, M.~Z.
\newblock \emph{The art of proving binomial identities}.
\newblock Chapman and Hall/CRC, 2019.

\bibitem[Stanley(1986)]{stanley1986enumerative}
Stanley, R.~P.
\newblock \emph{What is enumerative combinatorics?}
\newblock Springer, 1986.

\bibitem[Stein \& Joyner(2005)Stein and Joyner]{sage2005}
Stein, W. and Joyner, D.
\newblock Sage: system for algebra and geometry experimentation.
\newblock \emph{SIGSAM Bull.}, 39\penalty0 (2):\penalty0 61–64, June 2005.
\newblock ISSN 0163-5824.
\newblock \doi{10.1145/1101884.1101889}.
\newblock URL \url{https://doi.org/10.1145/1101884.1101889}.

\bibitem[Sturmfels(2002)]{sturmfels2002solving}
Sturmfels, B.
\newblock \emph{Solving systems of polynomial equations}.
\newblock Number~97. American Mathematical Soc., 2002.

\bibitem[Tsoukalas et~al.(2024)Tsoukalas, Lee, Jennings, Xin, Ding, Jennings, Thakur, and Chaudhuri]{tsoukalas2024putnambench}
Tsoukalas, G., Lee, J., Jennings, J., Xin, J., Ding, M., Jennings, M., Thakur, A., and Chaudhuri, S.
\newblock Putnambench: A multilingual competition-mathematics benchmark for formal theorem-proving.
\newblock In \emph{AI for Math Workshop@ ICML 2024}, 2024.

\bibitem[Wang et~al.(2025{\natexlab{a}})Wang, Unsal, Lin, Baksys, Liu, Santos, Sung, Vinyes, Ying, Zhu, et~al.]{wang2025kimina}
Wang, H., Unsal, M., Lin, X., Baksys, M., Liu, J., Santos, M.~D., Sung, F., Vinyes, M., Ying, Z., Zhu, Z., et~al.
\newblock Kimina-prover preview: Towards large formal reasoning models with reinforcement learning.
\newblock \emph{arXiv preprint arXiv:2504.11354}, 2025{\natexlab{a}}.

\bibitem[Wang et~al.(2024)Wang, Zhang, Jia, Pan, Diao, Pi, and Zhang]{wang2024theoremllama}
Wang, R., Zhang, J., Jia, Y., Pan, R., Diao, S., Pi, R., and Zhang, T.
\newblock Theoremllama: Transforming general-purpose llms into lean4 experts.
\newblock In \emph{2024 Conference on Empirical Methods in Natural Language Processing, EMNLP 2024}, pp.\  11953--11974. Association for Computational Linguistics (ACL), 2024.

\bibitem[Wang et~al.(2025{\natexlab{b}})Wang, Pan, Li, Zhang, Jia, Diao, Pi, Hu, and Zhang]{wang2025ma}
Wang, R., Pan, R., Li, Y., Zhang, J., Jia, Y., Diao, S., Pi, R., Hu, J., and Zhang, T.
\newblock Ma-lot: Model-collaboration lean-based long chain-of-thought reasoning enhances formal theorem proving.
\newblock \emph{arXiv preprint arXiv:2503.03205}, 2025{\natexlab{b}}.

\bibitem[Wang et~al.(2025{\natexlab{c}})Wang, Yao, Pan, Diao, and Zhang]{wang2025gar}
Wang, R., Yao, J., Pan, R., Diao, S., and Zhang, T.
\newblock Gar: Generative adversarial reinforcement learning for formal theorem proving, 2025{\natexlab{c}}.
\newblock URL \url{https://arxiv.org/abs/2510.11769}.

\bibitem[Wei et~al.(2024)Wei, Sun, and Wang]{wei2024proving}
Wei, C., Sun, M., and Wang, W.
\newblock Proving olympiad algebraic inequalities without human demonstrations.
\newblock \emph{arXiv preprint arXiv:2406.14219}, 2024.

\bibitem[Wilf(2005)]{wilf2005generatingfunctionology}
Wilf, H.~S.
\newblock \emph{generatingfunctionology}.
\newblock CRC press, 2005.

\bibitem[Wilf \& Zeilberger(1990)Wilf and Zeilberger]{1990WZHerbert}
Wilf, H.~S. and Zeilberger, D.
\newblock Rational functions certify combinatorial identities.
\newblock \emph{Journal of the American Mathematical Society}, 3\penalty0 (1):\penalty0 147--158, 1990.
\newblock ISSN 08940347, 10886834.
\newblock URL \url{http://www.jstor.org/stable/1990986}.

\bibitem[Wilf \& Zeilberger(1992)Wilf and Zeilberger]{wilf1992algorithmic}
Wilf, H.~S. and Zeilberger, D.
\newblock An algorithmic proof theory for hypergeometric (ordinary and “q”) multisum/integral identities.
\newblock \emph{Inventiones mathematicae}, 108\penalty0 (1):\penalty0 575--633, 1992.

\bibitem[Wu et~al.(2024)Wu, Huang, Zhou, Ying, Wang, Lin, and Chen]{InternLM2.5-StepProver}
Wu, Z., Huang, S., Zhou, Z., Ying, H., Wang, J., Lin, D., and Chen, K.
\newblock Internlm2.5-stepprover: Advancing automated theorem proving via expert iteration on large-scale {LEAN} problems.
\newblock \emph{CoRR}, abs/2410.15700, 2024.
\newblock \doi{10.48550/ARXIV.2410.15700}.
\newblock URL \url{https://doi.org/10.48550/arXiv.2410.15700}.

\bibitem[Xin et~al.(2024)Xin, Guo, Shao, Ren, Zhu, Liu, Ruan, Li, and Liang]{xin2024deepseek}
Xin, H., Guo, D., Shao, Z., Ren, Z., Zhu, Q., Liu, B., Ruan, C., Li, W., and Liang, X.
\newblock Deepseek-prover: Advancing theorem proving in llms through large-scale synthetic data.
\newblock \emph{arXiv preprint arXiv:2405.14333}, 2024.

\bibitem[Xin et~al.(2025)Xin, Xi, Yang, Chen, Wu, Xiao, Sun, Zheng, and Shen]{xin2025bfs}
Xin, R., Xi, C., Yang, J., Chen, F., Wu, H., Xiao, X., Sun, Y., Zheng, S., and Shen, K.
\newblock Bfs-prover: Scalable best-first tree search for llm-based automatic theorem proving.
\newblock \emph{arXiv preprint arXiv:2502.03438}, 2025.

\bibitem[Xiong et~al.(2025)Xiong, Lv, Shan, Wang, Yang, and Zhi]{xiong2025combinatorialidentitiesbenchmarktheorem}
Xiong, B., Lv, H., Shan, H., Wang, J., Yang, Z., and Zhi, L.
\newblock A combinatorial identities benchmark for theorem proving via automated theorem generation, 2025.
\newblock URL \url{https://arxiv.org/abs/2502.17840}.

\bibitem[Yang et~al.(2024)Yang, Swope, Gu, Chalamala, Song, Yu, Godil, Prenger, and Anandkumar]{Leandojo}
Yang, K., Swope, A., Gu, A., Chalamala, R., Song, P., Yu, S., Godil, S., Prenger, R.~J., and Anandkumar, A.
\newblock Leandojo: Theorem proving with retrieval-augmented language models.
\newblock \emph{Advances in Neural Information Processing Systems}, 36, 2024.

\bibitem[Yu et~al.(2025{\natexlab{a}})Yu, Zhang, Zhu, Yuan, Zuo, Yue, Dai, Fan, Liu, Liu, Liu, Lin, Lin, Ma, Sheng, Tong, Zhang, Zhang, Zhang, Zhu, Zhu, Chen, Chen, Wang, Yu, Song, Wei, Zhou, Liu, Ma, Zhang, Yan, Qiao, Wu, and Wang]{yu2025dapo}
Yu, Q., Zhang, Z., Zhu, R., Yuan, Y., Zuo, X., Yue, Y., Dai, W., Fan, T., Liu, G., Liu, L., Liu, X., Lin, H., Lin, Z., Ma, B., Sheng, G., Tong, Y., Zhang, C., Zhang, M., Zhang, W., Zhu, H., Zhu, J., Chen, J., Chen, J., Wang, C., Yu, H., Song, Y., Wei, X., Zhou, H., Liu, J., Ma, W.-Y., Zhang, Y.-Q., Yan, L., Qiao, M., Wu, Y., and Wang, M.
\newblock Dapo: An open-source llm reinforcement learning system at scale, 2025{\natexlab{a}}.
\newblock URL \url{https://arxiv.org/abs/2503.14476}.

\bibitem[Yu et~al.(2025{\natexlab{b}})Yu, Peng, Ding, Li, Peng, Liu, Zhang, Yuan, Xin, Huang, et~al.]{yu2025formalmath}
Yu, Z., Peng, R., Ding, K., Li, Y., Peng, Z., Liu, M., Zhang, Y., Yuan, Z., Xin, H., Huang, W., et~al.
\newblock Formalmath: Benchmarking formal mathematical reasoning of large language models.
\newblock \emph{arXiv preprint arXiv:2505.02735}, 2025{\natexlab{b}}.

\bibitem[Zeilberger(1991)]{zeilberger1991method}
Zeilberger, D.
\newblock The method of creative telescoping.
\newblock \emph{J. Symb. Comput.}, 11\penalty0 (3):\penalty0 195--204, 1991.

\bibitem[Zhou et~al.(2025)Zhou, Lei, Zhou, Sun, Zhu, Ye, Zhang, Liu, Wei, and Liu]{zhou2025spark}
Zhou, X., Lei, Y., Zhou, X., Sun, J., Zhu, Y., Ye, Z., Zhang, W., Liu, Q., Wei, S., and Liu, C.
\newblock Spark-prover-x1: Formal theorem proving through diverse data training.
\newblock \emph{arXiv preprint arXiv:2511.13043}, 2025.

\end{thebibliography}
 
\newpage
\appendix
\onecolumn
  
\section{Glossary of Symbolic and Formal-Proof Terms}
\label{appendix:term_chart}
To better understand the terms, we provide this chart that explains in detail every term, abbreviation, and
corresponding tool.

\subsection{WZ and Symbolic-Computation Terms}

\paragraph{Hypergeometric term.}
A bivariate term $F(n,k)$ is \emph{(bi-)hypergeometric} if the ratios
$F(n,k+1)/F(n,k)$ and $F(n+1,k)/F(n,k)$ are rational functions in $(n,k)$.
Many WZ/creative-telescoping procedures apply only to identities whose summands admit such rational ratios.

\paragraph{Telescoping sum.}
A sum of the form $\sum_k \bigl(G(n,k+1)-G(n,k)\bigr)$ is \emph{telescoping}:
all intermediate terms cancel, leaving only boundary contributions.
Thus, if the boundary terms vanish, the whole sum is evaluated to $0$.

\paragraph{WZ pair.}
A \emph{WZ pair} is a pair of terms $(F,G)$ that satisfy
\[
F(n+1,k)-F(n,k)=G(n,k+1)-G(n,k),
\]
so that summing over $k$ yields a (typically constant) behavior of
$S(n)=\sum_k F(n,k)$ under suitable boundary conditions.

\paragraph{WZ certificate $R(n,k)$.}
In WZ proofs one often sets $G(n,k)=R(n,k)F(n,k)$, where $R(n,k)$ is a rational function.
 This $R$ is called a \emph{certificate}; once proposed by a CAS, the remaining work is to verify the induced rational identity and the boundary conditions within the proof assistant.

\paragraph{Creative telescoping / Zeilberger algorithm.}
\emph{Creative telescoping} generalizes WZ by producing higher-order relations
\[
\sum_{j=0}^{J} a_j(n)\,F(n+j,k)=G(n,k+1)-G(n,k),
\]
which induce a linear recurrence for $S(n)=\sum_k F(n,k)$ after summing over $k$.
The case $J=1$ recovers the classical WZ equation.

\paragraph{Recurrence relation.}
A \emph{recurrence} is an equation that relates $S(n)$ to different $n$ (e.g., first-order or higher-order linear recurrences).
In our setting, proving the recurrence together with a base case is sufficient to determine $S(n)$ and close the target identity.

\paragraph{Boundary conditions / boundary obligations.}
To justify telescoping in a finite range (e.g., $k=0,\dots,n$), one must prove that the boundary terms vanish (or match) at the endpoints, such as $G(n,0)=0$ and $G(n,n+1)=0$.
These are emitted as explicit \emph{boundary obligations}.

\paragraph{Normalization.}
For an identity $\sum_k f(n,k)=r(n)$ with $r(n)\neq 0$, we normalize to
$\sum_k F(n,k)=1$ by setting $F(n,k)=f(n,k)/r(n)$.
This reduces the goal to proving the constancy of the normalized sum plus a base case.

\paragraph{Summation range transformation.}
Many identities require rewriting sums to a canonical range, e.g.,
\[
\sum_{k=a}^{b} f(k)=\sum_{k=0}^{b-a} f(k+a),
\]
Separate a few boundary terms.
Such transformations are often needed to match the requirements of WZ-style decomposition.

\paragraph{Singularity / nonzero side conditions.}
Symbolic certificates frequently introduce denominators.
Formal verification therefore requires \emph{side conditions} ensuring that those denominators are nonzero in the relevant domain (e.g., to justify \texttt{field\_simp} steps in Lean).

\subsection{Automated Theorem Proving Terms}

\paragraph{Kernel verification.}
Lean's trusted kernel checks that the produced proof term/script is valid.
All reported successes correspond to proofs accepted by the kernel.

\paragraph{Proof obligation.}
A \emph{proof obligation} is a subgoal generated by a proof sketch (e.g., a boundary condition or a recurrence lemma) that must be proved to complete the overall proof.

\paragraph{pass@k (e.g., pass@32).}
\texttt{pass@k} measures whether at least one of the $k$ independently sampled proof attempts verifies.
It is standard for stochastic theorem provers where success is probabilistic.

\paragraph{Executable proof sketch.}
In this paper, a \emph{proof sketch} is not a natural-language outline but a structured, machine-checkable decomposition (e.g., recurrence + boundary obligations + base case) that can be executed in Lean as a sequence of formal subgoals.

\paragraph{WZ-covered vs.\ WZ-uncovered identities.}
An identity is \emph{WZ-covered} if the symbolic engine can synthesize a WZ/creative-telescoping certificate and emit a sketch.
Otherwise, it is \emph{WZ-uncovered} and is treated as a direct proving target.

\paragraph{WZ sub-goals and CI proving tasks.}
\emph{WZ sub-goals} refer to the boundary obligations and recurrence lemmas produced by sketch construction.
\emph{CI proving tasks} include both these sub-goals and the WZ-uncovered identities, forming the unified task set solved by WZ-Prover.

\paragraph{Expert iteration / verifier-filtered bootstrapping.}
We repeatedly sample candidate proofs from the current prover for a pool of unproven statements, keep only Lean-verified proofs, and add the resulting pairs (statement, proof) back to the training set. We then update the prover and continue on the remaining statements, expanding coverage while maintaining formal correctness.

\paragraph{Goal / theorem statement.}
A \emph{goal} $G$ denotes a Lean proposition to be proved (either a full identity or an intermediate lemma).
We write $\mathcal{V}(G,\pi)=\texttt{true}$ if the Lean kernel verifies that the tactic script $\pi$ proves $G$.

\paragraph{Candidate statement.}
A \emph{candidate statement} is an unlabeled target identity extracted from external sources (e.g., tables of identities) and normalized into a uniform, Lean-parseable form.
The pool of candidate statements is used for expert-in-the-loop bootstrapping and data expansion.

\section{Pseudocode for WZ-LLM and WZ-Prover Training}
\label{sec:Pseudocode}
Our approach combines the complementary strengths of symbolic computation and learned proof search: symbolic WZ decomposition externalizes long-horizon proof planning into a set of locally checkable obligations, while a specialized Lean prover efficiently discharges the resulting algebraic and boundary subgoals.
To make this neuro-symbolic workflow precise and reproducible, we summarize the inference-time framework and the corresponding training pipeline in Algorithm~\ref{alg:wz-llm} and Algorithm~\ref{alg:wzprover_train}. 

We present our end-to-end proving framework \textsc{WZ-LLM} (Algorithm~\ref{alg:wz-llm}) and the training pipeline for the resulting prover WZ-Prover (Algorithm~\ref{alg:wzprover_train}).
At inference time, \textsc{WZ-LLM} queries a symbolic engine to construct a WZ-style proof sketch, consisting of a recurrence structure together with boundary-condition obligations.
When WZ decomposition succeeds, the framework reduces the original identity into a set of Lean-checkable subgoals, which are then discharged by WZ-Prover in Lean4.
If the identity is \emph{WZ-inapplicable} (i.e., no certificate/sketch can be synthesized), \textsc{WZ-LLM} falls back to direct proving using WZ-Prover.
All generated proofs and proof fragments are strictly verified by the Lean kernel.

\begin{algorithm}[tb]
\caption{\textsc{WZ-LLM}: Sketch-Guided Neuro-Symbolic Proving in Lean4}
\label{alg:wz-llm}
\begin{algorithmic}
\STATE {\bfseries Input:} target goal $S$, WZ engine $\mathcal{E}$, prover $\mathcal{M}$ (WZ-Prover), Lean verifier $\mathcal{V}$
\STATE {\bfseries Output:} Lean4-verified proof of $S$ or {\sc Fail}

\STATE $\mathcal{Q} \leftarrow \{S\}$ \COMMENT{queue of pending obligations}
\STATE $\mathcal{P} \leftarrow \emptyset$ \COMMENT{kernel-checkable proof fragments}
\STATE $\mathcal{S}_{\text{seen}} \leftarrow \emptyset$ \COMMENT{visited goals (dedup)}

\WHILE{$\mathcal{Q} \neq \emptyset$}
  \STATE Pop $G$ from $\mathcal{Q}$
  \IF{$G \in \mathcal{S}_{\text{seen}}$}
    \STATE {\bfseries continue}
  \ENDIF
  \STATE $\mathcal{S}_{\text{seen}} \leftarrow \mathcal{S}_{\text{seen}} \cup \{G\}$

  \STATE $(\texttt{applicable},\ \texttt{sketch},\ \mathcal{L}) \leftarrow \mathcal{E}(\textsc{WZ-Decompose},G)$

  \IF{\texttt{applicable}}
    \STATE $\mathcal{P} \leftarrow \mathcal{P} \cup \{\texttt{sketch}\}$
    \COMMENT{Lean skeleton reducing $G$ to obligations in $\mathcal{L}$}
    \FOR{each $\ell \in \mathcal{L}$}
      \IF{$\ell \notin \mathcal{S}_{\text{seen}}$}
        \STATE Push $\ell$ into $\mathcal{Q}$
      \ENDIF
    \ENDFOR
  \ELSE
    \STATE $\pi \leftarrow \mathcal{M}(G)$ \COMMENT{direct proving for WZ-inapplicable goals}
    \IF{$\mathcal{V}(G,\pi)$}
      \STATE $\mathcal{P} \leftarrow \mathcal{P} \cup \{\pi\}$
    \ELSE
      \STATE {\bfseries return} {\sc Fail}
    \ENDIF
  \ENDIF
\ENDWHILE

\STATE {\bfseries return} \textsc{Assemble}$(\mathcal{P})$
\COMMENT{assemble fragments; Lean kernel checks final proof}
\end{algorithmic}
\end{algorithm}

\paragraph{Algorithm~\ref{alg:wz-llm}: Sketch-guided neuro-symbolic proving.}
Algorithm~\ref{alg:wz-llm} describes the inference-time procedure of \textsc{WZ-LLM}. 
Given a target Lean goal $S$, the framework maintains a queue $\mathcal{Q}$ of pending obligations and iteratively processes each goal $G$ popped from the queue. 
To avoid redundant work (e.g., repeated lemmas produced by different decompositions), we maintain a visited set $\mathcal{S}_{\text{seen}}$ and skip any goal that has already been processed.

For each goal $G$, \textsc{WZ-LLM} queries the symbolic backend $\mathcal{E}$ via \textsc{WZ-Decompose}. 
If the goal is \emph{WZ-applicable}, the backend returns (i) an executable Lean \emph{sketch} and (ii) a finite set of Lean-checkable obligations $\mathcal{L}$, which typically include recurrence/ratio checks, boundary conditions, normalization steps, and non-vanishing side conditions needed to justify telescoping. 
The sketch is stored as a kernel-checkable fragment in $\mathcal{P}$, while each obligation $\ell\in\mathcal{L}$ is pushed into $\mathcal{Q}$ for subsequent discharge (subject to deduplication by $\mathcal{S}_{\text{seen}}$). 

If \textsc{WZ-Decompose} reports that $G$ is \emph{WZ-inapplicable}, the system falls back to direct proving: the learned prover $\mathcal{M}$ (WZ-Prover) generates a tactic script $\pi\leftarrow\mathcal{M}(G)$, which is accepted only if the Lean verifier $\mathcal{V}$ confirms $\mathcal{V}(G,\pi)=\texttt{true}$. 
Any verifier failure causes the procedure to return \textsc{Fail}. 
Finally, once $\mathcal{Q}$ is exhausted, \textsc{Assemble}$(\mathcal{P})$ combines all verified fragments (sketches and lemma proofs) into a complete end-to-end proof, which is again checked by the Lean kernel.

\paragraph{Algorithm~\ref{alg:wzprover_train}: Training pipeline for WZ-Prover.}
Algorithm~\ref{alg:wzprover_train} details the three-stage training pipeline that produces WZ-Prover. 
The procedure starts from a small expert-verified seed dataset $\mathcal{D}_{\text{seed}}$ and performs supervised fine-tuning (SFT) to obtain an initial prover $\mathcal{M}$. 
It then performs expert-in-the-loop dataset expansion for $T$ rounds over a pool of candidate goals $\mathcal{S}_{\text{cand}}$.
In each round $t$, the algorithm generates new training samples $\mathcal{D}_{\text{new}}$ from two sources. 
When a sampled goal $S$ is \emph{WZ-applicable}, we decompose it into obligations $\mathcal{L}$ and attempt to prove each lemma $\ell\in\mathcal{L}$ using the current prover $\mathcal{M}$. 
Draft proofs $\hat{\pi}_{\ell}$ are then curated through \textsc{ExpertRepairAndFilter}, which repairs incomplete or partially correct traces, filters out low-quality outputs, and returns a cleaned pair $(\ell',\pi_{\ell})$ only when a proof is well-formed and verifiable. 
When $S$ is \emph{WZ-inapplicable}, we instead attempt to directly prove the full identity with $\mathcal{M}$ and apply the same expert repair/filtering step, yielding $(S',\pi_S)$ when successful. 
All accepted samples must pass Lean verification under $\mathcal{V}$.

After collecting $\mathcal{D}_{\text{new}}$, we augment the training set $\mathcal{D}\leftarrow \mathcal{D}\cup\mathcal{D}_{\text{new}}$ and retrain the prover with SFT on the expanded Lean-verified dataset, repeating for $T$ rounds. 
Finally, we apply DAPO refinement to further improve long-horizon lemma chaining and robustness. 
In contrast to modifying the reward design, our refinement relies on \emph{difficulty-smoothing data curation}: we run a rollout-based diagnostic pass to estimate goal pass rates under the current policy, bucket goals into coarse difficulty bins, and then downsample redundant easy instances while removing near-zero-pass-rate goals that contribute mostly noisy gradients (Appendix~\ref{app:dapo_refine}). 
DAPO is then applied on this curated RL set using the same reward definition as in Eq.~\ref{eq:dapo_total_reward_final}. 
The resulting model $\mathcal{M}^{\star}$ is returned as the final WZ-Prover.

\begin{algorithm}[tb]
  \caption{WZ-Prover Training: Seed SFT $\rightarrow$ Expert Iteration $\rightarrow$ DAPO Refinement}
  \label{alg:wzprover_train}
  \begin{algorithmic}
    \STATE {\bfseries Input:} seed proofs $\mathcal{D}_{\text{seed}}$, candidate goals $\mathcal{S}_{\text{cand}}$, WZ engine $\mathcal{E}$, Lean verifier $\mathcal{V}$, expert rounds $T$
    \STATE {\bfseries Output:} trained prover $\mathcal{M}^{\star}$ (WZ-Prover)

    \STATE $\mathcal{M} \leftarrow \textsc{SFT}(\mathcal{D}_{\text{seed}})$ \hfill // cold-start
    \STATE $\mathcal{D} \leftarrow \mathcal{D}_{\text{seed}}$

    \FOR{$t = 1$ {\bfseries to} $T$}
      \STATE $\mathcal{D}_{\text{new}} \leftarrow \emptyset$

      \FOR{{\bfseries each} sampled goal $S \in \mathcal{S}_{\text{cand}}$}
        \STATE $(\texttt{applicable},\ \texttt{sketch},\ \mathcal{L}) \leftarrow \mathcal{E}(\textsc{WZ-Decompose}, S)$

        \IF{\texttt{applicable}}
          \FOR{{\bfseries each} lemma $\ell \in \mathcal{L}$}
            \STATE $\hat{\pi}_{\ell} \leftarrow \mathcal{M}(\ell)$
            \STATE $(\ell', \pi_{\ell}) \leftarrow \textsc{ExpertRepairAndFilter}(\ell, \hat{\pi}_{\ell})$
            \IF{$(\ell', \pi_{\ell}) \neq \emptyset$ {\bfseries and} $\mathcal{V}(\ell', \pi_{\ell})$}
              \STATE $\mathcal{D}_{\text{new}} \leftarrow \mathcal{D}_{\text{new}} \cup \{(\ell', \pi_{\ell})\}$
            \ENDIF
          \ENDFOR
        \ELSE
          \STATE $\hat{\pi}_{S} \leftarrow \mathcal{M}(S)$
          \STATE $(S', \pi_{S}) \leftarrow \textsc{ExpertRepairAndFilter}(S, \hat{\pi}_{S})$
          \IF{$(S', \pi_{S}) \neq \emptyset$ {\bfseries and} $\mathcal{V}(S', \pi_{S})$}
            \STATE $\mathcal{D}_{\text{new}} \leftarrow \mathcal{D}_{\text{new}} \cup \{(S', \pi_{S})\}$
          \ENDIF
        \ENDIF
      \ENDFOR

      \STATE $\mathcal{D} \leftarrow \mathcal{D} \cup \mathcal{D}_{\text{new}}$
      \STATE $\mathcal{M} \leftarrow \textsc{SFT}(\mathcal{D})$ \hfill // retrain on expanded data
    \ENDFOR

  
        \STATE // DAPO refinement with difficulty-smoothing data curation (Appendix~\ref{app:dapo_refine})
    \STATE $\mathcal{M}^{\star} \leftarrow \textsc{DAPO}(\mathcal{M};\ R(\pi;G))$
  \STATE {\bfseries return} $\mathcal{M}^{\star}$
  \end{algorithmic}
\end{algorithm}

\section{More Experimental Details}
\label{app:exp_details}
This appendix reports implementation and reproducibility details, covering additional experimental settings (tooling, token budgets, and optimization hyperparameters).

\subsection{Software Stack and Tooling}
\label{app:stack}

\noindent
\textbf{Lean environment.}
All proofs are checked in \texttt{leanprover/lean4:v4.25.0}.
Kernel verification is treated as the sole correctness criterion.

\noindent
\textbf{Symbolic backend.}
We use SageMath \texttt{10.7} for symbolic computation, including WZ certificate/recurrence synthesis and auxiliary ratio simplifications.
Sage is invoked programmatically via Python.

\noindent
\textbf{Inference runtime.}
Model inference is run with vLLM \texttt{0.13.0}.
For baseline provers that rely on long chain-of-thought (Kimina-Prover, MA-LoT, DeepSeek-Prover-V2, and Goedel-Prover-V2), we cap the reasoning length at \textbf{16,384} tokens to reduce compute while keeping the sampling budget comparable (pass@32 where applicable).
Other baselines follow the official decoding and token-limit settings provided in their released code.

\subsection{Evaluation Budget and Token Limits.}
We report end-to-end success under pass@32 where applicable.
To keep compute manageable for long-reasoning baselines (Kimina-Prover, MA-LoT, DeepSeek-Prover-V2, Goedel-Prover-V2), we cap the reasoning length at \textbf{16,384} tokens.
Other baselines follow their official evaluation settings. 

\subsection{Prompting Protocol}
\label{app:prompts}

We use the same instruction-following structure throughout training and evaluation.
Each instance asks the model to (i) provide a short proof plan that identifies key lemmas/tactics, and (ii) output a complete Lean4 proof script that closes the goal without using \texttt{sorry}.
For WZ-induced obligations, the prompt additionally includes CAS-produced artifacts (e.g., rational certificates or simplified ratios) as explicit context, so that Lean-side verification focuses on well-definedness, algebraic rewriting, and recurrence/telescoping checks.

\section{Training Data Details}
\label{app:data_format}

Our training corpus contains two major categories:
(i) \emph{symbolic-induced lemmas} produced by WZ-style decomposition (and related algebraic simplifications), and
(ii) \emph{end-to-end identity proofs} for goals that can be discharged directly (including identities not covered by WZ decomposition).
All retained samples must be compile and verified in Lean; we also duplicate the proofs before adding them to the training set.

\noindent
\textbf{Expert-verified iteration.}
We perform expert-verified dataset expansion for $T=2$ rounds.
In each round, we sample identities from the candidate pool, apply symbolic decomposition to generate lemma obligations, and let the prover attempt both (i) lemma proofs for sketch-induced subgoals and (ii) direct end-to-end identity proofs.
We retain only samples that compile and verify in Lean, and we deduplicate accepted proofs before adding them to the training corpus in Table~\ref{tab:data_expansion}.

\noindent
\textbf{RL dataset size and composition.}
The RL stage operates on a filtered subset of the Lean-verified SFT corpus obtained from the seed set plus expert-verified iteration (Table~\ref{tab:data_expansion}).
To smooth the difficulty mismatch between sketch-induced lemmas (often short and repetitive) and full-identity goals (typically long-horizon and failure-prone), we run a rollout-based diagnostic pass and estimate a pass rate for each goal under the current policy.
We then construct the RL set by downsampling large clusters of high-pass-rate, near-duplicate easy instances and excluding near-zero-pass-rate goals whose rollouts are dominated by truncation or early failure.
The resulting RL set emphasizes diverse mid-to-hard goals, contains a higher fraction of long-horizon proof trajectories, and still mixes lemma-style subgoals (e.g., ratio/non-vanishing obligations) with full identity goals to preserve end-to-end capability.
All RL prompts follow the same prover input template, and each rollout is scored by Lean kernel verification plus the soft overlong punishment in Eq.~\ref{eq:dapo_total_reward_final}.

\begin{table}[t]
\centering
\small
\setlength{\tabcolsep}{6pt}
\begin{tabular}{lrrrr}
\toprule
\textbf{Stage} &
\textbf{Accepted identities} &
\textbf{Seed lemmas} &
\textbf{Generated obligations} &
\textbf{Accepted lemmas} \\
\midrule
Seed dataset
& 307
& 1{,}200
& --
& 1{,}200 \\
\midrule
Expert iteration (Round~1)
& 32
& --
& 7{,}385
& 5{,}139 \\
Expert iteration (Round~2)
& 79
& --
& 2{,}246
& 532 \\
\midrule
\textbf{Total (accepted)}
& \textbf{418}
& --
& --
& \textbf{6{,}871} \\
\bottomrule
\end{tabular}
\caption{\textbf{Training data growth via expert-verified iteration.}
We start from a seed corpus of 307 Lean-verified identities and 1{,}200 framework-extracted lemma proofs.
Each iteration samples identity statements, decomposes them into lemma obligations when applicable, and retains only Lean-verified and deduplicated outputs.
Across two rounds, expert iteration adds 111 new identity proofs and 5{,}671 lemma proofs, yielding 418 identities and 6{,}871 lemmas in total.}
\label{tab:data_expansion}
\end{table}

\subsection{Lemma types and data format}
Symbolic-induced training instances are dominated by two recurring lemma families: (i) \textbf{ratio lemmas}, which expose CAS-precomputed term ratios (e.g., A(n,k{+}1)/A(n,k) or A(n{+}1,k)/A(n,k)) to avoid expensive normalization inside Lean; and (ii) \textbf{non-vanishing (well-definedness) lemmas}, which establish that denominators are nonzero, enabling safe use of \texttt{field\_simp} and preventing singularities during normalization and recurrence checking.
All lemma samples are stored as instruction-following triples (\texttt{instruction}/\texttt{input}/\texttt{output}): the \texttt{input} contains a Lean goal template, and the \texttt{output} includes (a) a brief proof plan and (b) a complete Lean4 proof script.
We give one example for each lemma type below (formatted identically to training instances).

The first example illustrates a typical ``local obligation'' that frequently appears in our pipeline
(e.g., when normalizing expressions or discharging boundary side conditions). Such goals are not
mathematically deep, but they are essential for making later \texttt{simp}/\texttt{field\_simp}
steps well-defined and for keeping the main proof trace stable.

\lstset{
  basicstyle=\ttfamily\small,
  breaklines=true,
  breakatwhitespace=true,
  columns=fullflexible,
  keepspaces=true,
  showstringspaces=false
}

\begin{tcolorbox}[
  colback=gray!6,
  colframe=black!25,
  boxrule=0.5pt,
  arc=2pt,
  left=6pt,right=6pt,top=6pt,bottom=6pt
]
\small

\noindent\textbf{Statement (Lean goal).}
\begin{lstlisting}[language=lean]
import Mathlib

open Nat Finset BigOperators

theorem factorial_succ_ge_one {n : ℕ} (h : 1 ≤ n) :
  n * (n - 1)! = n! := by
  -- to be completed
\end{lstlisting}

\medskip
\noindent\textbf{Proof / Model Output.}

\noindent\textbf{Problem.}
Show that for any natural number $n$ with $1 \le n$, we have
\[
n \cdot (n-1)! = n!.
\]

\medskip
\noindent\textbf{Key observations.}
\begin{enumerate}
  \item The hypothesis $1 \le n$ implies $0 < n$, which lets us rewrite $n$ as $(n-1)+1$ via \texttt{Nat.succ\_pred\_eq\_of\_pos}.
  \item Lean's factorial recurrence is encoded as \texttt{Nat.factorial\_succ}: \texttt{(m+1)! = (m+1) * m!}.
        Therefore, once we rewrite $n$ as \texttt{(n-1)+1}, the goal reduces to a direct rewrite by \texttt{Nat.factorial\_succ}.
  \item No induction is needed: the proof is a one-step normalization that bridges the arithmetic identity \texttt{n = (n-1)+1} with the definitional equation of factorial.
\end{enumerate}

\medskip
\noindent\textbf{Proof plan.}
\begin{tcolorbox}[
  colback=white,
  colframe=black!15,
  boxrule=0.4pt,
  arc=2pt,
  left=6pt,right=6pt,top=6pt,bottom=6pt
]
\small
\begin{enumerate}
  \item Derive \texttt{0 < n} from the assumption \texttt{1 $\leq$ n}.
  \item Rewrite \texttt{n} as \texttt{(n - 1) + 1} using \texttt{Nat.succ\_pred\_eq\_of\_pos}.
  \item Replace \texttt{n!} by \texttt{((n-1)+1)!} and rewrite it using \texttt{Nat.factorial\_succ}.
  \item Finish by simplification (\texttt{simp}) and associativity/commutativity normalization (\texttt{ring} is unnecessary for $\mathbb{N}$ here).
\end{enumerate}
\end{tcolorbox}

\medskip
\noindent\textbf{Complete Lean 4 Proof.}
\begin{lstlisting}[language=lean]
import Mathlib

open Nat Finset BigOperators

theorem factorial_succ_ge_one {n : ℕ} (h : 1 ≤ n) :
  n * (n - 1)! = n! := by
  have hn : 0 < n := Nat.pos_of_ne_zero (by
    -- 1 ≤ n implies n ≠ 0
    exact Nat.ne_of_gt (Nat.lt_of_lt_of_le Nat.zero_lt_one h))
  -- rewrite n as (n - 1) + 1
  have hn' : (n - 1).succ = n := Nat.succ_pred_eq_of_pos hn
  -- turn it into a rewriting form n = (n - 1) + 1
  have hn'' : n = (n - 1) + 1 := by
    -- succ is +1 in Nat
    simpa [Nat.succ_eq_add_one] using hn'.symm
  -- use factorial_succ on m = n-1
  -- (n-1+1)! = (n-1+1) * (n-1)!
  -- then rewrite (n-1+1) as n
  calc
    n * (n - 1)! = ((n - 1) + 1) * (n - 1)! := by
      simpa [hn''] 
    _ = ((n - 1) + 1)! := by
      simpa [Nat.factorial_succ]
    _ = n! := by
      simpa [hn'']
\end{lstlisting}

\end{tcolorbox}

The second example illustrates a typical \emph{non-vanishing / well-definedness} obligation produced by symbolic preprocessing:
from an arithmetic assumption $n>m$ in $\mathbb{N}$, the prover must show that a real denominator term $(n-m+1)$ is nonzero.

\lstset{
  basicstyle=\ttfamily\small,
  breaklines=true,
  breakatwhitespace=true,
  columns=fullflexible,
  keepspaces=true,
  showstringspaces=false
}

\begin{tcolorbox}[
  colback=gray!6,
  colframe=black!25,
  boxrule=0.5pt,
  arc=2pt,
  left=6pt,right=6pt,top=6pt,bottom=6pt
]
\small

\noindent\textbf{Statement (Lean goal).}
\begin{lstlisting}[language=lean]
import Mathlib

open Real Nat Finset BigOperators Polynomial

set_option maxHeartbeats 8000000000
theorem hwz (n m : ℕ) (htotalNumidx : n > m) :
  ((↑n - ↑m + 1) : ℝ) ≠ 0 := by
  -- to be completed
\end{lstlisting}

\medskip
\noindent\textbf{Proof / Model Output.}

\noindent\textbf{Problem.}
Given natural numbers $n,m$ with $n>m$, prove the real expression
\[
(n - m + 1) \neq 0.
\]

\medskip
\noindent\textbf{Key observations.}
\begin{enumerate}
  \item From $n>m$ we get $m < n+1$ in $\mathbb{N}$ immediately. This is a cleaner arithmetic target than reasoning directly about a casted subtraction in $\mathbb{R}$.
  \item If we can show $(m : \mathbb{R}) < (n+1 : \mathbb{R})$, then
  \[
  (n - m + 1 : \mathbb{R}) = (n+1 : \mathbb{R}) - (m : \mathbb{R}) > 0,
  \]
  hence it cannot be zero (discharged by \texttt{linarith}).
  \item The bridge between the $\mathbb{N}$-inequality and the $\mathbb{R}$-inequality is handled by \texttt{norm\_cast}. The remaining linear arithmetic on naturals is solved by \texttt{omega}.
\end{enumerate}

\medskip
\noindent\textbf{Proof plan.}
\begin{tcolorbox}[
  colback=white,
  colframe=black!15,
  boxrule=0.4pt,
  arc=2pt,
  left=6pt,right=6pt,top=6pt,bottom=6pt
]
\small
\begin{enumerate}
  \item Reduce the non-vanishing goal to a positivity/inequality statement:
        it suffices to prove $(m : \mathbb{R}) < (n+1 : \mathbb{R})$, then conclude by \texttt{linarith}.
  \item Use \texttt{norm\_cast} to turn $(m : \mathbb{R}) < (n+1 : \mathbb{R})$ into the natural-number inequality $m < n+1$.
  \item Prove $m < n+1$ from the hypothesis $n>m$ using \texttt{omega}.
\end{enumerate}
\end{tcolorbox}

\medskip
\noindent\textbf{Complete Lean 4 Proof.}
\begin{lstlisting}[language=lean]
import Mathlib

open Real Nat Finset BigOperators Polynomial

set_option maxHeartbeats 8000000000
theorem hwz (n m : ℕ) (htotalNumidx : n > m) :
  ((↑n - ↑m + 1) : ℝ) ≠ 0 := by
  -- Reduce to a strict inequality in ℝ, then finish by linear arithmetic.
  suffices h : (m : ℝ) < (n + 1 : ℝ) by
    linarith
  -- Move the goal back to ℕ.
  norm_cast
  -- Discharge the ℕ inequality from n > m.
  omega
\end{lstlisting}

\end{tcolorbox}




\subsection{Qualitative examples of WZ-Prover outputs}
\label{app:qual_examples}

We conclude this section with representative WZ-Prover outputs, illustrating the two complementary roles played by the trained prover in our framework.

\paragraph{Prompting protocol.}
WZ-Prover is queried with a fixed instruction-following prompt that asks the model to (i) first write a brief proof plan, and (ii) then output a complete Lean 4 proof script.
Concretely, we use the following prompt template, where \texttt{ts} is a placeholder for the current Lean proof state (the program replaces \texttt{ts} with the target theorem statement and its surrounding Lean context before inference):

\begin{tcolorbox}[
  colback=gray!6,
  colframe=black!25,
  boxrule=0.5pt,
  arc=2pt,
  left=6pt,right=6pt,top=6pt,bottom=6pt
]
\small
\noindent\textbf{Prompt template (\texttt{ts} is replaced at runtime).}

Complete the following Lean 4 code:

```lean4
ts```

Before producing the Lean 4 code to formally prove the given theorem, provide a detailed proof plan outlining the main proof steps and strategies.
The plan should highlight key ideas, intermediate lemmas, and proof structures that will guide the construction of the final formal proof.

\end{tcolorbox}

\medskip
\noindent
Using this prompt, we show qualitative examples of the resulting model outputs. These examples illustrate the two complementary roles played by the trained prover in our framework.

\begin{tcolorbox}[
  colback=gray!6,
  colframe=black!25,
  boxrule=0.5pt,
  arc=2pt,
  left=6pt,right=6pt,top=6pt,bottom=6pt
]
\small
\noindent\textbf{Two output modes.}
\begin{itemize} 
  \item \textbf{(A) Sketch-induced lemmas (symbolic-guided).}
  When the symbolic backend succeeds, a WZ sketch decomposes the original identity into many short Lean-checkable obligations.
  Among the most frequent are:
  \textit{(i) term-ratio recurrences} (used to verify the WZ invariant and telescoping), and
  \textit{(ii) non-vanishing denominator goals} (well-definedness conditions required by \texttt{field\_simp} and normalization).

  \item \textbf{(B) Direct end-to-end proofs (symbolic-free).}
  For WZ-inapplicable identities or when decomposition fails, WZ-Prover attempts to prove the identity directly in Lean, without relying on the symbolic sketch.
\end{itemize}
\end{tcolorbox}

\lstset{
  basicstyle=\ttfamily\small,
  breaklines=true,
  breakatwhitespace=true,
  columns=fullflexible,
  keepspaces=true,
  showstringspaces=false
}

\paragraph{(A) Example: a sketch-induced lemma.}
The following are typical lemma forms produced by symbolic decomposition.
They are short, local, and appear repeatedly across WZ-style proof chains.

\begin{tcolorbox}[
  colback=white,
  colframe=black!15,
  boxrule=0.4pt,
  arc=2pt,
  left=6pt,right=6pt,top=6pt,bottom=6pt
]
\small
\noindent\textbf{(A1) Term-ratio recurrence (for simplification).}
\begin{lstlisting}[language=lean]
import Mathlib
open Real Nat Finset BigOperators

set_option maxHeartbeats 8000000000
theorem rwz (n x : ℕ) (hby_cases : x ≥ n) :
    ((-1) ^ (n+1) : ℝ) = ((-1) ^ n) * (-1) := by
  -- typical local rewrite lemma used in ratio simplification
  simpa [pow_succ] using (pow_succ (-1 : ℝ) n)
\end{lstlisting}
\end{tcolorbox}

\begin{tcolorbox} [
  colback=white,
  colframe=black!15,
  boxrule=0.4pt,
  arc=2pt,
  left=6pt,right=6pt,top=6pt,bottom=6pt
]
\medskip
\noindent\textbf{(A2) Non-vanishing denominator (for \texttt{field\_simp}).}
\begin{lstlisting}[language=lean]
import Mathlib
open Real Nat Finset BigOperators

set_option maxHeartbeats 8000000000
theorem hwz (n m : ℕ) (h : n > m) : ((↑n - ↑m + 1) : ℝ) ≠ 0 := by
  -- typical well-definedness obligation: denominator minimal factor ≠ 0
  have : (0 : ℝ) < (↑n - ↑m + 1) := by
    have : (m : ℝ) < (n + 1 : ℝ) := by
      norm_cast; omega
    linarith
  linarith
\end{lstlisting}
\end{tcolorbox}

\paragraph{(B) Example: a direct identity proof (symbolic-free).}
In contrast, the next example is an end-to-end identity proof generated by WZ-Prover without invoking WZ decomposition.
It showcases a ``global'' proof that relies on Mathlib's analytic lemmas rather than sketch-induced local obligations.

\begin{tcolorbox}[
  colback=white,
  colframe=black!15,
  boxrule=0.4pt,
  arc=2pt,
  left=6pt,right=6pt,top=6pt,bottom=6pt
]
\small
\begin{lstlisting}[language=lean]
import Mathlib
open Nat BigOperators

theorem idt_149 (n : ℕ) {x : ℝ} (hx : 0 ≤ x ∧ x < 1) :
    (1 - x) ^ (-(n + 1 : ℕ) : ℝ) = ∑' k : ℕ, (n + k).choose k * x ^ k := by
  have hx_abs : |x| < 1 := by
    have : -1 < x ∧ x < 1 := by
      constructor
      · linarith [hx.1]
      · exact hx.2
    simpa [abs_lt] using this

  rw [neg_eq_neg_one_mul, Real.rpow_mul, Real.rpow_neg_one, inv_eq_one_div]
  norm_cast
  rw [Real.rpow_natCast, one_pow, div_eq_mul_inv]
  -- main analytic lemma (name may vary slightly across Mathlib versions)
  simpa using (tsum_choose_mul_geometric_of_abs_lt_1 (n := n) hx_abs)
\end{lstlisting}
\end{tcolorbox}

\noindent
Together, (A) and (B) reflect the intended complementarity:
symbolic decomposition turns long identities into many short obligations that WZ-Prover can discharge reliably,
while direct proving covers WZ-inapplicable identities without requiring a symbolic sketch.

\section{Case Studies}
\label{case}

We present two complementary case studies to illustrate how \textsc{WZ-LLM} operates in practice.
Appendix~\ref{case1} focuses on the \emph{symbolic proof sketch} produced by our framework, highlighting how symbolic computation induces a clear, executable global proof plan.
Appendix~\ref{case2} then demonstrates how this sketch is instantiated into a complete end-to-end Lean4 proof through a single interactive command, showcasing the full automation pipeline of \textsc{WZ-LLM} (Algorithm~\ref{alg:wz-llm}).

\subsection{From a Proof Sketch Template to Structured Obligations}
\label{case1}

This case study illustrates how \textsc{WZ-LLM} bridges symbolic proof planning and verifier-checked formal reasoning in Lean4. Given a target combinatorial identity, the framework first constructs a high-level \emph{WZ proof sketch template} following the standard Wilf--Zeilberger reasoning pattern: normalization, recurrence/telescoping, and boundary-condition discharge.
Rather than attempting unconstrained end-to-end proof search, this sketch explicitly exposes the global proof structure and reduces the original goal to a small set of well-scoped, Lean-checkable proof obligations.

\medskip
\noindent
The template below summarizes the canonical structure of a WZ-style proof sketch generated by \textsc{WZ-LLM}.
Each step corresponds to a distinct class of obligations that can be independently verified in Lean, allowing scalable and modular formalization.
\begin{tcolorbox}[
  colback=gray!3,
  colframe=black!35,
  boxrule=0.5pt,
  arc=2pt,
  title=\textbf{WZ Proof Sketch Template in Lean 4}
]
\small
\noindent
\textbf{Annotation convention.}
Subgoals delegated to WZ-Prover are marked as
\texttt{[LLM task: non-vanishing \& ratio lemmas]}.
These obligations typically establish well-definedness conditions
(nonzero denominators required by \texttt{field\_simp})
and verify CAS-derived term-ratio identities used in recurrence checking.

\medskip
\begin{lstlisting}
theorem <IdentityName> (n : ℕ) (hn : <premises>) :
 (∑ k in Finset.range (n + 1), A n k) = B n := by
\end{lstlisting}

\noindent
\textbf{Step 1: WZ certificate synthesis (CAS).}\\
Obtain a rational WZ certificate $R(n,k)$ from a symbolic backend (e.g., Sage).
\begin{lstlisting}
 let R : ℕ → ℕ → ℝ := WZ Certificate from CAS 
\end{lstlisting}

\noindent
\textbf{Step 2: Side conditions (well-definedness and boundaries).}\\
Prove non-vanishing conditions required for normalization and algebraic rewriting.
\begin{lstlisting}
have ne_zeroA : ∀ n k, k < n → (A n k : ℝ) ≠ 0 := by
-- [LLM task: non-vanishing \& ratio lemmas]}
have ne_zeroB : ∀ n, (B n : ℝ) ≠ 0 := by
-- [LLM task: non-vanishing \& ratio lemmas]}
\end{lstlisting}

\noindent
\textbf{Step 3: Normalization.
Reduce $\sum A(n,k)=B(n)$ to the canonical form $\sum F(n,k)=1$ using a fixed lemma.}\\

\begin{lstlisting}
have WZ_aux (n : ℕ) (f : ℕ → ℕ → ℝ) (B : ℕ → ℝ)
(ne_zero : ∀ n, (B n : ℝ) ≠ 0) :
(∑ k in Finset.range (n+1), (f n k / B n : ℝ) = 1)
   ↔ (∑ k in Finset.range (n+1), f n k = B n) := by
-- fixed normalization lemma}
\end{lstlisting}

\noindent
\textbf{Step 4: CAS-assisted ratio lemmas.}\\
Precompute quotient identities to simplify recurrence checking in Lean.
\begin{lstlisting}
have ratio_k : ∀ n k, k < n → A n (k+1)/A n k = <CAS_ratio_k> := by
-- [LLM task: non-vanishing \& ratio lemmas]}
have ratio_n : ∀ n k, k < n → A (n+1) k/A n k = <CAS_ratio_n> := by
-- [LLM task: non-vanishing \& ratio lemmas]}
have ratio_B : ∀ n, B (n+1)/B n = <CAS_ratio_B> := by
-- [LLM task: non-vanishing \& ratio lemmas]}
\end{lstlisting}

\noindent
\textbf{Step 5: WZ invariant.}\\
Verify the core WZ relation
$F(n\!+\!1,k)-F(n,k)=G(n,k\!+\!1)-G(n,k)$.
\begin{lstlisting}
let F := fun n k => A n k / B n
let G := fun n k => R n k * F n k
have WZ_invariant : ∀ n k, k < n →
F (n+1) k - F n k = G n (k+1) - G n k := by
-- [LLM task: non-vanishing \& ratio lemmas]}
\end{lstlisting}

\noindent
\textbf{Step 6: Telescoping and boundary discharge.}\\
Sum the invariant over $k$ to obtain $f(n+1)-f(n)=0$.
\begin{lstlisting}
let f := fun n => ∑ k in Finset.range (n+1), F n k
have telescoping_step : ∀ n, f (n+1) - f n = 0 := by
-- [LLM task: non-vanishing \& ratio lemmas]}
\end{lstlisting}

\noindent
\textbf{Step 7: Constant extraction and unnormalization.}\\
Prove the base case, conclude $f(n)=1$, and recover the original identity.
\end{tcolorbox}

\medskip
\noindent
This template makes explicit which parts of the proof are handled symbolically, which reduce to algebraic side conditions, and which require Lean-level reasoning.
Crucially, many boundary and non-vanishing conditions that are traditionally discharged manually in proof assistants are surfaced as explicit obligations and automatically handled by WZ-Prover.

\subsection{One-Command End-to-End Formal Proof in Lean}
\label{case2}

Beyond exposing a structured proof sketch, \textsc{WZ-LLM} is integrated into Lean as an interactive tactic.
For a concrete identity such as
\[
\sum_{k \in \mathrm{range}(n+1)} (-1)^k \binom{n}{k}\,\frac{m}{m+k}
= \frac{1}{\binom{m+n}{n}},
\]
the user only needs to invoke a single command:
\begin{center}
\textbf{\texttt{wz\_prove}}
\end{center}

\begin{figure*}[htp!]
    \centering
    \includegraphics[width=0.8\textwidth]{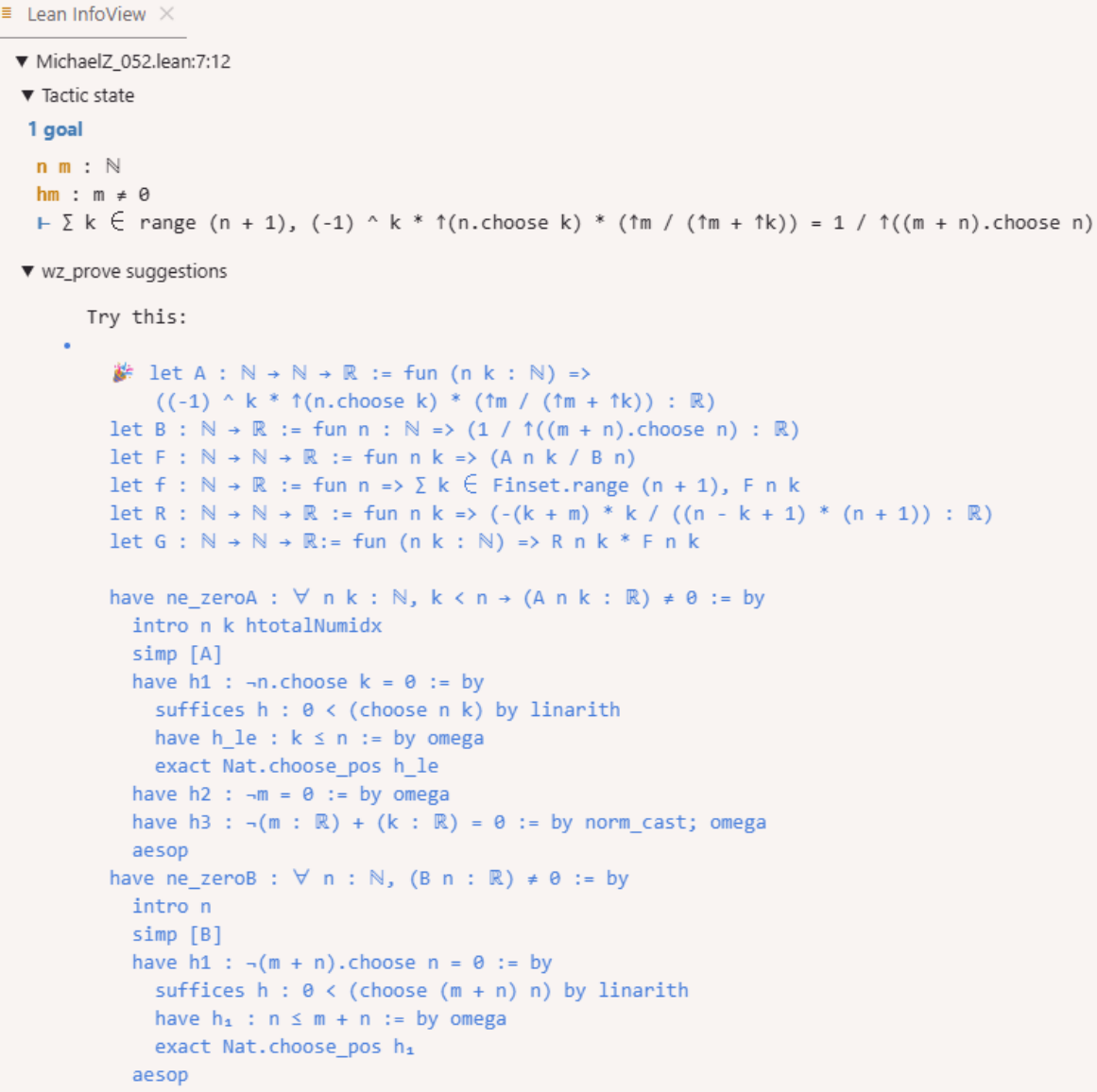}
    \caption{Lean InfoView output of the \textbf{\texttt{wz\_prove}} tactic. The tactic automatically invokes symbolic computation and the trained prover to generate Lean proof suggestions, which can be accepted to fill in the complete formal proof.}
    \label{fig:wzllm_leantactic}
\end{figure*}

\noindent
As illustrated in Figure~\ref{fig:wzllm_leantactic}, the Lean InfoView then automatically orchestrates the entire pipeline:
it calls a symbolic engine (Sage) to synthesize WZ certificates and recurrence relations, decomposes the goal into structured subgoals, and interacts with WZ-Prover to generate Lean proof scripts for each obligation.
Once the suggested proof is accepted in the InfoView, the complete proof is inserted into the Lean file, as shown in Listing 1, and verified by the kernel, yielding a fully checked end-to-end formal proof with minimal user intervention.

\begin{lstlisting}[caption={Complete Lean4 proof of the combinatorial identity $MichaelZ\_052$ generated by WZ-LLM.}]
import Mathlib 
open Real Nat Finset BigOperators Polynomial

set_option maxHeartbeats 8000000000
theorem MichaelZ_052 (n m : ℕ) {m : m ≠ 0} :
    ∑ k ∈ Finset.range (n + 1),
        (-1 : ℝ) ^ k * (Nat.choose n k) * (m / (m + k))
      = 1 / (Nat.choose (m + n) n : ℝ) := by

    --------------------------------------------------------------------
    -- (0) WZ-style normalization: define summand A(n,k), target B(n),
    --     normalized term F(n,k) := A(n,k)/B(n), and sum f(n)=∑F(n,k).
    --     Also define a certificate R(n,k) and its companion G(n,k)=R*F.
    --------------------------------------------------------------------
    let A : ℕ → ℕ → ℝ := fun (n k : ℕ) =>
        ((-1) ^ k * ↑(n.choose k) * (↑m / (↑m + ↑k)) : ℝ)
    let B : ℕ → ℝ := fun n : ℕ => (1 / ↑((m + n).choose n) : ℝ)
    let F : ℕ → ℕ → ℝ := fun n k => (A n k / B n)
    let f : ℕ → ℝ := fun n => ∑ k ∈ Finset.range (n+1), F n k

    -- WZ certificate (precomputed by symbolic backend in the full system)
    let R : ℕ → ℕ → ℝ := fun n k => (-(k + m)*k/((n - k + 1)*(n + 1)) : ℝ)
    let G : ℕ → ℕ → ℝ:= fun (n k : ℕ) => R n k * F n k

    --------------------------------------------------------------------
    -- (1) Non-vanishing side conditions: needed for field_simp/division.
    --     These are typical well-definedness lemmas in WZ pipelines.
    --------------------------------------------------------------------
    have ne_zeroA : ∀ n k : ℕ, k < n → (A n k : ℝ) ≠ 0 := by
      intro n k htotalNumidx
      simp [A]
      -- choose n k ≠ 0 when k ≤ n (here k < n)
      have h1 : ¬n.choose k = 0 := by
        suffices h : 0 < (choose n k) by linarith
        have h_le : k ≤ n := by omega
        exact Nat.choose_pos h_le
      -- m ≠ 0 (from implicit parameter), plus denominator m+k ≠ 0
      have h2 : ¬m = 0 := by omega
      have h3 : ¬(m : ℝ)+ (k : ℝ)= 0 := by norm_cast; omega
      aesop
    have ne_zeroB : ∀ n : ℕ, (B n : ℝ)  ≠ 0 := by
      intro n
      simp [B]
      -- choose (m+n) n ≠ 0 since n ≤ m+n
      have h1 : ¬(m + n).choose n = 0 := by
        suffices h : 0 < (choose (m + n) n) by linarith
        have h₁ : n ≤ m + n := by omega
        exact Nat.choose_pos h₁
      aesop
    have ne_zeroB_succ : ∀ n : ℕ, (B (n + 1) : ℝ)  ≠ 0 := by
      intro n
      simp [B]
      aesop
    have ne_zeroB': ∀ n : ℕ, (B n : ℝ) * (B (n + 1) : ℝ)  ≠ 0 := by
      intro m
      rw [mul_ne_zero_iff]
      exact And.intro (ne_zeroB m) (ne_zeroB (m + 1))
    have ne_zeroAB' : ∀ n k : ℕ, k < n → (B n : ℝ) * (A n k : ℝ) ≠ 0 := by
      intro n k hidx
      rw [mul_ne_zero_iff]
      exact And.intro (ne_zeroB n) (ne_zeroA n k hidx)

    --------------------------------------------------------------------
    -- (2) Algebraic factoring helpers: isolate ratios like
    --     A(n+1,k)/A(n,k) and B(n+1)/B(n) after field_simp.
    --     These are bookkeeping lemmas that keep later steps readable.
    --------------------------------------------------------------------
    have l₁ : ∀ n k : ℕ, k < n
        → A (n + 1) k * B n / (B n * A n k) = A (n + 1) k / A n k := by
      intro n k hidx
      field_simp [ne_zeroA, ne_zeroB]

    have l₂ : ∀ n k : ℕ, k < n
        →  B (n + 1) * A n k / (B n * A n k) =  B (n + 1) / B n := by
      intro n k hidx
      field_simp [ne_zeroA, ne_zeroB]

    have r₁ : ∀ n k : ℕ, k < n → B (n + 1) * (R n (k + 1) * A n (k + 1))
        / (B n * A n k) = R n (k + 1) * (A n (k + 1) / A n k) * (B (n + 1) / B n) := by
      intro n k hidx
      field_simp [ne_zeroA, ne_zeroB]

    have r₂ : ∀ n k : ℕ, k < n → B (n + 1) * (A n k * R n k)
        / (B n * A n k) = R n k * (B (n + 1) / B n) := by
      intro n k hidx
      field_simp [ne_zeroA, ne_zeroB]

    --------------------------------------------------------------------
    -- (3) WZ normalization lemma: sum (A/B) = 1  <->  sum A = B.
    --     This bridges the normalized constant-sum formulation back to
    --     the original target identity at the very end.
    --------------------------------------------------------------------
    have WZ_aux (n : ℕ) (f : ℕ → ℕ → ℝ) (B : ℕ → ℝ)
        (ne_zero : ∀ n : ℕ, (B n : ℝ) ≠ 0) :
        ∑ k ∈ Finset.range (n+1), (f n k / B n : ℝ) = (1 : ℝ)
        ↔ ∑ k ∈ Finset.range (n+1), f n k = B n := by
      constructor
      · intro h
        rw [← Finset.sum_div, div_eq_iff (ne_zero n), one_mul] at h
        norm_cast at h
      · intro _
        rw [← Finset.sum_div, div_eq_iff (ne_zero n), one_mul]
        norm_cast

    have Step1 := WZ_aux n A B ne_zeroB

    --------------------------------------------------------------------
    -- (4) Symbolic (ratio) facts used by the WZ identity:
    --     aux₁: k-shift ratio   A(n,k+1)/A(n,k)
    --     aux₂: n-shift ratio   A(n+1,k)/A(n,k)
    --     aux₃: target ratio    B(n+1)/B(n)
    --     These are exactly the kind of lemmas that a CAS-backed stage
    --     tends to generate in bulk for many problems.
    --------------------------------------------------------------------
    have aux₁ (n k : ℕ) (htotalNumidx : k < n):
        A n (k + 1) / A n k = -(n - k)*(k + m)/((k + m + 1)*(k + 1)) := by
      simp only [A]
      -- (-1)^(k+1) = (-1)^k * (-1)
      have r1 : ((-1) ^ (k + 1) : ℝ) = ((-1) ^ k) * (-1) := by
        rw [pow_succ]
      rw [r1]
      -- choose ratio: C(n,k+1) expressed from C(n,k)
      have r2 : (↑(n.choose (k + 1)) : ℝ) = (↑(n.choose k)) * (n - k)/(k + 1) := by
        have h₁ : n ≥ k := by omega
        have h₂ : (k + 1 : ℝ) ≠ 0 := by linarith
        field_simp
        norm_cast
        simp [Nat.choose_succ_right_eq]
      rw [r2]
      -- side conditions for field_simp
      have h1 : (↑(n.choose k) : ℝ) ≠ 0 := by
        norm_cast
        suffices h : 0 < (n.choose k) by exact h.ne'
        apply Nat.choose_pos
        omega
      have h2 : ((↑m + ↑(k + 1)) : ℝ) ≠ 0 := by norm_cast; omega
      have h3 : ((↑m + ↑k) : ℝ) ≠ 0 := by norm_cast; omega
      have h4 : (((-1) ^ k) : ℝ) ≠ 0 := by norm_num
      have h5 : ((↑k + 1) : ℝ) ≠ 0 := by norm_cast
      have h6 : (↑m : ℝ) ≠ 0 := by norm_cast
      have h7 : ((↑k + ↑m + 1) : ℝ) ≠ 0 := by norm_cast
      field_simp
      grind

    have aux₂ (n k : ℕ) (htotalNumidx : k < n):
        A (n + 1) k / A n k = (n + 1)/(n - k + 1) := by
      simp only [A]
      -- choose lifting: C(n+1,k) expressed from C(n,k)
      have r1 : (↑((n + 1).choose k) : ℝ) = (↑(n.choose k)) * (n + 1)/(n - k + 1) := by
        have h₁ : n ≥ k := by omega
        have h₂ : (↑n - ↑k + 1 : ℝ) ≠ 0 := by norm_cast
        field_simp
        norm_cast
        simp [Nat.choose_mul_succ_eq]
        omega
      rw [r1]
      -- side conditions for field_simp
      have h1 : (↑(n.choose k) : ℝ) ≠ 0 := by
        norm_cast
        suffices h : 0 < (n.choose k) by exact h.ne'
        apply Nat.choose_pos
        omega
      have h2 : ((↑m + ↑k) : ℝ) ≠ 0 := by norm_cast; omega
      have h3 : (((-1) ^ k) : ℝ) ≠ 0 := by norm_num
      have h4 : (↑m : ℝ) ≠ 0 := by norm_cast
      have h5 : ((↑n - ↑k + 1) : ℝ) ≠ 0 := by
        suffices 0 < (↑n - ↑k + 1 : ℝ) by linarith
        simp only [sub_eq_add_neg, add_assoc]
        norm_cast
        omega
      field_simp

    have aux₃ : ∀ n : ℕ, B (n + 1) / B n = (n + 1)/(n + m + 1) := by
      intro n
      simp only [B]
      -- ratio of binomial coefficients in B(n)=1/choose(m+n,n)
      have r1 : (↑((m + (n + 1)).choose (n + 1)) : ℝ)
            = (↑((m + n).choose n)) * (n + m + 1)/(n + 1) := by
        have h₁ : (n + 1 : ℝ) ≠ 0 := by linarith
        field_simp
        norm_cast
        have h₂ : (m + (n + 1)).choose (n + 1) * (n + 1)
              = (m + n + 1).choose (n + 1) * (n + 1) := by
          have h₂₁ : m + (n + 1) = m + n + 1 := by omega
          rw [h₂₁]
        have h₂' : (m + n).choose n * (n + m + 1)
              = (m + n).choose n * (m + n + 1) := by
          have h₂'₁ : n + m + 1 = m + n + 1 := by omega
          rw [h₂'₁]
        simp [h₂, h₂', Nat.choose_succ_right_eq, Nat.choose_mul_succ_eq]
      rw [r1]
      -- side conditions
      have h1 : ((↑n + ↑m + 1) : ℝ) ≠ 0 := by norm_cast
      have h2 : (1 : ℝ) ≠ 0 := by norm_num
      have h3 : ((↑n + 1) : ℝ) ≠ 0 := by norm_cast
      have h4 : (↑((m + n).choose n) : ℝ) ≠ 0 := by
        norm_cast
        push_neg
        rw [← Nat.pos_iff_ne_zero]
        apply Nat.choose_pos
        linarith
      field_simp

    -- boundary helpers that appear in the telescoping boundary terms
    have r_aux1 : ∀ n : ℕ, A (n + 1) (n+1) = ((-1)*(m + n) / (m + n + 1)) * A n n := by
      intro n
      simp only [A]
      field_simp
      have r1 : ((-1) ^ (n + 1) : ℝ) = ((-1) ^ n) * (-1) := by rw [pow_succ]
      rw [r1]
      norm_num
      grind

    have r_aux2 : ∀ n : ℕ, A (n + 1) n = (n + 1) * A n n := by
      intro n
      simp only [A]
      field_simp
      norm_num

    --------------------------------------------------------------------
    -- (5) Core WZ step: show f(n+1) - f(n) = 0, by telescoping:
    --     F(n+1,k) - F(n,k) = G(n,k+1) - G(n,k)
    --     Sum over k collapses to boundary terms, which cancel.
    --------------------------------------------------------------------
    have Step2 : ∀ n : ℕ, f (n + 1) - f n = 0 := by
      intro n

      -- pointwise WZ identity for each k (local recurrence + certificate)
      have WZ (k : ℕ) (htotalNumidx:k < n) :
          F (n + 1) k - F n k = G n (k + 1) - G n k := by
        simp only [F, G]
        -- normalize differences under common denominators
        field_simp [ne_zeroA, ne_zeroB]
        rw [← div_left_inj' (ne_zeroAB' n k (by omega))]
        rw [sub_div, mul_sub, sub_div ]
        -- split into ratios A(n+1,k)/A(n,k) and B(n+1)/B(n)
        rw [l₁ n k (by omega), l₂ n k (by omega), r₁ n k (by omega), r₂ n k (by omega)]
        -- insert the explicit symbolic ratios aux₁/aux₂/aux₃
        rw [sub_eq_iff_eq_add, ← sub_mul]
        nth_rw 2 [show B (n + 1) / B n = 1 * (B (n + 1) / B n) by grind]
        rw [← add_mul]
        simp [R]
        rw [aux₁ n k (by linarith), aux₂ n k (by linarith), aux₃ n]
        -- finish by algebraic simplification (denominators shown nonzero)
        field_simp
        have h1 : ((↑n - (↑k + 1) + 1) : ℝ) ≠ 0 := by
          suffices h : k < (n : ℝ) by linarith
          norm_cast
        have h2 : ((↑k + 1) : ℝ) ≠ 0 := by norm_cast
        have h3 : ((↑n - ↑k + 1) : ℝ) ≠ 0 := by
          suffices h : k < (n + 1 : ℝ) by linarith
          norm_cast
          omega
        have h4 : ((↑k + ↑m + 1) : ℝ) ≠ 0 := by norm_cast
        have h5 : ((↑n + 1) : ℝ) ≠ 0 := by norm_cast
        field_simp
        ring

      -- sum the pointwise WZ identity over k and telescope G(n,·)
      calc f (n + 1) - f n
        _ = (∑ k ∈ range (n+1), (F (n + 1) k - F n k)) + F (n + 1) (n+1) := by
          simp [f]
          rw [Finset.sum_range_add]
          simp only [range_one, sum_singleton, add_zero, sub_add_eq_add_sub]
        _ = (∑ k ∈ range n, (G n (k + 1) - G n k)) + F (n + 1) n - F n n + F (n + 1) (n+1) := by
          rw [Finset.sum_range_add]
          simp only [range_one, sum_singleton, add_zero, add_left_inj, add_sub]
          congr 2
          apply Finset.sum_congr rfl
          intro k hidx
          simp only [mem_range] at hidx
          exact WZ k hidx
        _ = (G n n - G n 0) + F (n + 1) n - F n n + F (n + 1) (n+1) := by
          -- telescoping: ∑ (G(k+1)-G(k)) = G(n)-G(0)
          congr 3
          apply sum_range_sub

      -- discharge boundary terms and show total difference is zero
      simp [G, F]
      field_simp [ne_zeroB n, ne_zeroB_succ n, ne_zeroB' n]
      simp [mul_comm (B n)]
      rw [add_sub_right_comm, ← sub_mul, add_assoc, ← add_mul]
      nth_rw 1 [show B (n + 1) = (B (n + 1) / B n) * B n by field_simp [ne_zeroB]]
      conv_lhs => enter[1];rw[← mul_assoc]
      rw [← add_mul]
      simp only [_root_.mul_eq_zero]
      left
      rw [sub_right_comm,← sub_one_mul,aux₃ n, r_aux1 n, r_aux2 n]
      simp only [A, R]
      field_simp
      norm_num
      grind

    --------------------------------------------------------------------
    -- (6) Conclude f(n) is constant (=1) by Step2 and base case f(0)=1.
    --------------------------------------------------------------------
    have Step3 : ∀ n : ℕ, f n = 1 := by
      intro n
      induction' n with n hm
      · -- base case: f(0)=F(0,0)=A(0,0)/B(0)=1
        simp [f, F, A, B]
        have h1 : (↑m : ℝ) ≠ 0 := by norm_cast
        field_simp
      · -- inductive step: f(n+1)=f(n) from Step2, then use IH
        exact (sub_eq_zero.1 $ Step2 n).trans (hm)

    --------------------------------------------------------------------
    -- (7) Unnormalize back to the original statement: ∑ A = B.
    --------------------------------------------------------------------
    unfold A B at Step1
    rw [Step1.1]
    exact Step3 n


   
\end{lstlisting}

\end{document}